\documentclass[11pt]{article}

\usepackage{header}

\def\kvn{Kvetsh}
\def\moss{Untangled Lariat}

\makeatletter
\renewcommand*{\@fnsymbol}[1]{\ensuremath{
  \ifcase#1\or {\, \scriptstyle \spadesuit} \or {\, \scriptstyle \blacklozenge} \or {\, \scriptstyle \clubsuit}
    \or \mathsection\or \mathparagraph\or \|\or **\or \dagger\dagger
   \or \ddagger\ddagger \else\@ctrerr\fi}}
\makeatother

\title{Untangling Lariats: \\
Subgradient Following of Variationally Penalized Objectives}

\author{
  Kai-Chia Mo\thanks{\,\! \sf{qwe7859126@gmail.com}}
  \and
  Shai Shalev-Shwartz\thanks{\,\! Hebrew University, \sf{shais@cs.huji.ac.il}}
  \and
  Nis$\ae$l Sh\'artov\thanks{\,\! Nis$\ae$l Consulting, \sf{nisael@sonic.net}}}

\date{}

\begin{document}
\maketitle

\begin{abstract}
We describe an apparatus for subgradient-following of the optimum of convex
problems with variational penalties. In this setting, we receive a sequence
$y_i,\ldots,y_n$ and seek a smooth sequence $x_1,\ldots,x_n$. The smooth
sequence needs to attain the minimum Bregman divergence to an input sequence
with additive variational penalties in the general form of
$\sum_i{}g_i(x_{i+1}-x_i)$. We derive known algorithms such as the fused lasso
and isotonic regression as special cases of our approach. Our
approach also facilitates new variational penalties such as non-smooth barrier
functions.

We then derive a novel lattice-based procedure for
subgradient following of variational penalties characterized through the output
of arbitrary convolutional filters. This paradigm yields efficient solvers for
high-order filtering problems of temporal sequences in which sparse discrete
derivatives such as acceleration and jerk are desirable.
We also introduce and analyze new multivariate problems in which
$\x_i,\y_i\in\reals^d$ with variational penalties that depend on
$\|\x_{i+1}-\x_i\|$. The norms we consider are $\ell_2$ and $\ell_\infty$ which
promote group sparsity.
\end{abstract}

\section{Sub-Introduction}
To start, let us examine two seemingly different problems, the fused
lasso~\cite{tibshirani2005sparsity, FriedmanHaHoTi07} and isotonic
regression~\cite{AyerBrEwReSi55, BarlowUb71}. The input to the two procedures
is a sequence $\y=y_1,\ldots,y_n$ where the goal is two obtain smoothed
sequences, denoted by $\x$, that are the solutions of the following
optimization problems,
\begin{eqnarray*}
(\text{Fused Lasso}) &
  \displaystyle \min_{\x\in\reals^n} &\!\!\!
    \frac12\|\x-\y\|^2 + \sum_{i=1}^n \lambda_i |x_{i+1} - x_i| \\
(\text{Iso. Entropy Regression}) &
  \displaystyle \min_{\x\in[0,1]^n} &\!\!\!
  \sum_{i=1}^n x_i \log\frac{x_i}{y_i} + (1-x_i) \log\frac{1-x_i}{1-y_i}
    ~\mbox{ s.t. }~ \forall i\in[n\!-\!1]:\, x_{i+1} \geq x_{i}
    ~~.
\end{eqnarray*}
We can unify the two problems into a single abstract problem of the form,
\beq{generic}
\min_{\x}\:
  \sum_{i=1}^n h_i(x_i) + \sum_{i=1}^{n-1} g_i(x_i - x_{i+1}) ~~.
\eeq

Here, both $h_i:\reals\to\reals_+$ and $g_i:\reals\to\reals_+$ are convex in
their single argument. Specific choices for $h_i$ and $g_i$ yield as special
cases the fused-lasso, by setting $ g_i(x_i - x_{i+1}) = \lambda_i |x_i - x_{i+1}|$, and
isotonic regression, by defining $ g_i(x_i - x_{i+1}) = 0$ if $x_{i+1} \geq
x_i$ and $\infty$ otherwise. Albeit the superficially different form of the
variational penalty $g_i$, our apparatus facilitates gradient following
procedures which differ by a {\em single} line of code as shown in Fig.~8.  As
in the case of isotonic regression, this procedure can facilitate non-smooth
variational penalties for $g_i$. We fondly refer to the general
subgradient-following paradigm as the {\em \moss}. To illustrate the power of
the \moss\ we derive a {\em novel} derivative of the generic problem given
by \eqr{generic} which uses fairly general barrier functions for $g_i(\cdot)$.

We next introduce a construction from signal processing into
our setting by penalizing the convolution of $\x$ with a predefined convolution
filter $\bl$. Concretely, we provide an efficient procedure for finding the
optimum of,
$$
  \min_\x \sum_{i} h_i(x_i) + \sum_{i} \big|(\bl \ast \x)_{i+1}\big| ~~,
$$
where $\bl\in\reals^k$ is a predefined yet arbitrary filter of $k-1$ free
parameters. We introduce a subgradient-following procedure which operates over a
$k$-dimensional lattice. It takes, in the worst case, $\oo(n^k)$ operations to
construct the lattice from which the optimal solution is obtained.  In
applications where $\y$ represents a temporal sequence, the variational penalty
of the fused lasso $|x_{i+1}-x_i|$ promotes smoothed solutions with regions of
zero discrete ``velocity''. Analogously, the high-order \moss\ is a filtering
and smoothing apparatus for problems in which we seek regions of zero
acceleration, zero jerk, and other forms of finite differences. 

Last but not least, we generalize the \moss\ to {\em multivariate} settings in
which each element of the input sequence is a vector. We thus seek a solution
$\x_1, \ldots, \x_n$ where $\x_i\in\reals^d$ for all $i\in[n]$. In order to
obtain variational sparsity in the multivariate case we replace the absolute
value $|x_{i+1}-x_i|$ with norm penalties over the differences
$\|\x_{i+1}-\x_i\|$.  To start, we derive a closed-form for the squared
$2$-norm, $\|\x_{i+1}-\x_i\|^2$. We use the closed form solution to build
surrogate functions for the $2$-norm itself, $\|\x_{i+1}-\x_i\|$, and the
infinity-norm, $\|\x_{i+1}-\x_i\|_\infty$. The result is an iterative yet
efficient solver for the problem,
$$
  \min_{\x_1,\ldots,\x_n}\: \sum_{i=1}^n h_i(\x_i) +
  \sum_{i=1}^{n-1} \|\x_i - \x_{i+1}\|_p
  ~\mbox{ where }~ p\in\{2,\infty\} .$$
For brevity of the derivation of the multivariate setting we confine ourselves
to the case where $h_i(\x_i) = \|\x_i - \y_i\|^2$.

The initial motivation for this work can be traced back to the influential work
on the fused lasso~\cite{tibshirani2005sparsity, FriedmanHaHoTi07} and isotonic
regression~\cite{AyerBrEwReSi55, BarlowUb71, boyarshinov2006linear}. The setting
of the fused lasso has been revisited numerous times and found various venues
for applications. See for instance~\cite{hochbaum2001efficient,
hoefling2010path}, though any attempt to be exhaustive here is going to do
injustice to the plethora of applications that employ the fused lasso. Several
authors~\cite{davies2001local, johnson2013dynamic, kolmogorov2016total}
described and analyzed a linear time algorithm for solving the fused lasso
problem which is morally equivalent to \moss\ with the aforementioned specific
choices for $h_i$ and $g_i$.  Alternative algorithms for the fused lasso and
similar problems, such as the ones described in \cite{barbero2014modular,
condat2013direct}, do not guarantee a runtime linear in $n$ nonetheless
demonstrate excellent empirical results.

The paper that is morally most relevant to our work
is~\cite{kolmogorov2016total} where the authors follow the derivation
in~\cite{johnson2013dynamic} to generalize the algorithm to non-smooth unary
functions and to trees. The procedure described in~\cite{kolmogorov2016total} is
an instance of a primal-dual message passing algorithm~\cite{hazan2010norm}.
The authors also consider unary functions subject to a pairwise penalty term
which also include the fused lasso penalty as a special case. The lens of the
work presented in~\cite{kolmogorov2016total} is probabilistic in nature and does
not consider general pairwise penalties.

Several other forms of variational penalties were the focus specific extension
of the fused lasso and isotonic regression. For instance, the works presented
in~\cite{boyarshinov2006linear} considers isotonic regression with $\ell_1$
norm for the variational penalty. We readily obtain the same procedure as a
special case of our approach. The problem of high-order variational penalties
was originally proposed by~\cite{kim2009ell_1} and its statistical properties
were studied by~\cite{steidl2006splines, tibshirani2014adaptive}. An
algorithmic perspective was provided in~\cite{kim2009ell_1} which discusses an
approximate solution using a primal-dual interior point method. For penalties
of the form $\lambda \sum_i |x_{i+1} - x_i|$, the entire solution path for
admissible values of $\lambda$, is provided using a dual path algorithm in
\cite{tibshirani2011solution, arnold2016efficient}. To the best of our
knowledge, we give the first efficient {\em analytic} solution for a given
$\lambda$.

We generalize the aforementioned research papers by providing a single unified
algorithmic paradigm for general convex unary functions and a broad class of
pairwise terms. Our analysis technique deviates from prior research by
employing the Fenchel dual of the primal problem, a view that was advocated in
the context of online learning by Shalev-Shwartz~\cite{Shalev07}. For a
thorough study of Fenchel duality and its connection to Lagrange dual see
for instance Chapter 3 in~\cite{BorweinLewis06}. The incorporation of Fenchel
duality leads to a simple and useful generalization to Bregman
divergences~\cite{bregman1967relaxation} between $\x$ and $\y$ as the
objective. This generalization does not bear an additional computation cost.
We provide in an appendix Python code that implements subgradient following
procedures for most of the settings discussed in this paper.

\section{Abstract Subgradient Following Algorithm} \label{sbgf:sec}
Let us now consider a general minimization with sequential penalties
of the form,
\beq{prox2}
U = \min_{\x}\: \sum_{i=1}^n h_i(x_i)
	+ \sum_{i=1}^{n-1} g_i(x_i - x_{i+1}) ~~.
\eeq

Throughout the paper we assume that $h_i$ and $g_i$ are convex and lower
semi-continuous. The Fenchel dual of a function $f$ is defined as,
$$f^*(\alpha) = \sup_{x\in\mathrm{dom}(f)} \big\{\alpha x - f(x)\big\}~~.$$ The domain of the functions we consider is closed, compact, and convex. We can thus
replace the supremum above with a maximum. We next utilize Fenchel-Moreau
theorem for each $g_i$,
$$g_i(x_i - x_{i-1}) = \max_\alpha \big\{\alpha (x_i - x_{i-1}) - g_i^*(\alpha)\big\}~~,$$
in order to evaluate $U$ as follows,
\begin{eqnarray*}
	U & = & \min_{\x}\, \max_{\ba}\: \sum_{i=1}^n h_i(x_i)
	+ \sum_{i=1}^{n-1} \left[ \alpha_i(x_i - x_{i+1})
		- g_i^*(\alpha_i) \right] ~~,
\end{eqnarray*}
We confine ourselves to settings for which strong duality holds thus,
\begin{eqnarray*}
U & = & \min_{x_1}\, \max_{\alpha_1}\, \min_{x_2}
	\ldots \max_{\alpha_{n-1}}\, \min_{x_n}\: \sum_{i=1}^n h_i(x_i)
	+ \sum_{i=1}^{n-1} \left[ \alpha_i(x_i - x_{i+1})
		- g_i^*(\alpha_i) \right] ~~.
\end{eqnarray*}
To find the optimum we define a sequence of partial single variable
functions defined as,
\begin{eqnarray} \label{fixi}
f_i(x_i) & \equiv & \max_{\alpha_i}\,
	\ldots \min_{x_n}\: \sum_{j=i}^{n-1}
	\left[
		h_{j+1}(x_{j+1})
		+ \alpha_j(x_j - x_{j+1})
		- g_j^*(\alpha_j)
	\right] ~~.
\end{eqnarray}
For boundary conditions we let $\forall x: f_n(x) = 0$.
In addition, we define an auxiliary function
$f_{i-\ha}(x_i) = f_i(x_i) + h_i(x_i)$. Using these definitions we have
$U = \min_{x_1} f_\ha(x_1)$.
Next we introduce a summation-free form for $f_i$,
\begin{eqnarray}
f_i(x_i) & = & \max_{\alpha_i}\: \alpha_i x_i - g_i^*(\alpha_i)
	+ \min_{x_{i+1}} \left\{f_{i+1}(x_{i+1}) + h_{i+1}(x_{i+1})
	- \alpha_i x_{i+1} \right\} \nonumber \\
& = & \max_{\alpha_i}\: \alpha_i x_i - g_i^*(\alpha_i)
	+ \min_{x_{i+1}} \left\{f_{i+\ha}(x_{i+1})
	- \alpha_i x_{i+1} \right\} \nonumber \\
& = & \max_{\alpha_i}\:
  \alpha_i x_i - g_i^*(\alpha_i) - f_{i+\ha}^*(\alpha_i) ~~.
  \label{eqn:fgfs}
\end{eqnarray}

Note that to find $f_i(x_i)$ we need to obtain the minimum of
$f_{i+\ha}(x_{i+1}) - \alpha_i x_{i+1}$ with respect to $x_{i+1}$.  Using
this abstraction, obtaining the optimum for $x_{i+1}$ ``only'' requires the
knowledge of $\alpha_i$. Analogously, to calculate $f_i$ itself we need to find
the optimum of $\alpha{}x_i-g_i^*(\alpha)-f_{i+\ha}^*(\alpha)$ with
respect to $\alpha$ which gives us $\alpha_i$. This in turn only requires the
knowledge of $x_i$. To recap, the optimization sub-problem for $x_{i+1}$ solely
uses the information on $\alpha_i$ and analogously the optimization sub-problem
for $\alpha_{i}$ only requires information of $x_i$. Let
$\ix_1,\ia_1,\ix_2,\ia_2,\ldots,\ia_{n-1},\ix_n$ denote a sequence of {\em
locally-admissible} pairs $\ix_i,\ia_i$ and $\ia_i,\ix_{i+1}$ where the full
sequence is not necessarily globally optimal. The subgradient conditions for
pair-restricted optimality define two admissible sets,
\begin{eqnarray}
S_{i+\ha} & = &
  \big\{ (\ix_{i+1}, \ia_i) : \ia_i \in \partial
  f_{i+\ha}(\ix_{i+1}) \big\} \label{Sih} \\
S_i & = & \big\{ (\ix_i, \ia_i)
	: \ia_i \in \partial f_i(\ix_i) \big\} ~~. \label{Si}
\end{eqnarray}

Given
$S_{i+1} = \left\{ (\ix_{i+1}, \ia_{i+1})
	: \ia_{i+1}\in \partial f_{i+1}(\ix_{i+1}) \right\}$ we perform an expansion,
\begin{eqnarray*}
S_{i+\ha} & = & \big\{(\ix_{i+1}, \ia_i)
	: \ia_i\in \partial f_{i+\ha}(\ix_{i+1})\big\} \\
	& = & \big\{ (\ix_{i+1}, \ia_i)
		: \ia_i \in \partial f_{i+1}(\ix_{i+1})
			+ \partial h_{i+1}(\ix_{i+1}) \big\} \\
	& = & \big\{ (\ix_{i+1}, \ia_i)
		: \ia_i - \ia_{i+1} \in \partial h_{i+1}(\ix_{i+1})
			\;;\; (\ix_{i+1}, \ia_{i+1}) \in S_{i+1} \big\} ~~.
\end{eqnarray*}

We next derive a recursive map $S_{i+1} \mapsto S_i$ by going through
$S_{i+\ha}$. From the representation of $f_i(x_i)$ as given by
Eq.~\eqref{eqn:fgfs} we get that the pair $(\ix_i,\ia_i)$ is admissible when
$$
  \ia_i \in \partial f_i(\ix_i) ~\Leftrightarrow~
	\ix_i \in \partial g_i^*(\ia_i) + \partial f_{i+\ha}^*(\ia_{i}) ~~.
$$
Starting with the definition of $S_i$ and expanding it based on
the above property we get,
\begin{eqnarray*}
S_i
  & = & \big\{ (\ix_i, \ia_i) :
    \ia_i\in \partial f_i(\ix_i) \big\} \\
	& = & \big\{ (\ix_i, \ia_i) :
    \ix_i \in \partial g_i^*(\ia_i)
			+ \partial {f_{i+\ha}}^* (\ia_i) \big\} \\
	& = & \big\{ (\ix_i, \ia_i) :
		\ix_i - \ix_{i+1} \in \partial g_i^*(\ia_i) \;;\;
		(\ix_{i+1}, \ia_i) \in S_{i+\ha} \big\} ~~.
\end{eqnarray*}
For the last equality we used the fact that,
$$ (\ix_{i+1}, \ia_i) \in S_{i+\ha} ~\Leftrightarrow~
	 \ia_i \in \partial f_{i+\ha}(\ix_{i+1}) ~\Leftrightarrow~
	 \ix_{i+1} \in \partial f^*_{i+\ha}(\ia_i) ~~.
$$
In summary, we obtain the following aesthetically succinct recursion, 
$$
\begin{array}{lcl}
S_{i+\ha} & = & \left\{ (\ix, \ia)
	: \ia - \ia' \in \partial h_{i+1}(\ix)\;;\;
		(\ix, \ia') \in S_{i+1} \right\} \\
S_i & = & \left\{ (\ix, \ia)
	: \ix - \ix' \in \partial g_i^*(\ia)\;;\;
		(\ix', \ia) \in S_{i+\ha} \right\} ~~.
\end{array}
$$
To kick-start the recursive procedure we set $S_n$ as a boundary condition,
\begin{eqnarray*}
  S_n \equiv \big\{ (\ix, 0) : \ix \in \text{dom}(h_n) \big\} ~~.
\end{eqnarray*}
We end the recursion upon reaching $S_\ha$ and then set $x_1$ to be the left
argument of the pair $(\ix,0) \in S_\ha$. Once $x_1$ is inferred
we can read out the rest of the sequence, $x_i$ for $i>1$, by unravelling
the recurrence,
\begin{eqnarray*}
  \alpha_i &:=& \ia \, \mbox{ s.t. } (x_i, \ia) \in S_i \\
  x_{i+1}  &:=& \ix \, \mbox{ s.t. } (\ix, \alpha_i) \in S_{i+\ha} ~~.
\end{eqnarray*}

\section{Derived Algorithms} \label{dalgo}
In this section we apply the subgradient following framework to specific
settings. We first show that the fused lasso is a simple instance. We next
derive other variational (fused) penalties, including barrier penalties that
were not analyzed before. We also show that the squared distance between
$\x$ and $\y$ can be generalized at no additional computational cost to
separable Bregman divergences.

\subsection{Revisiting the Fused Lasso} \label{rereflasso}
To recover the Fused Lasso algorithm using the subgradient-following method we
choose, $h_i(x) = \frac{1}{2} (x - y_i)^2$ and $ g_i(\delta) = \lambda_i |\delta|$.
Using basic calculus these choices imply that
$\nabla h_i(x) = x - y_i$ and $g_i^*(\alpha) = \1_{|\alpha|\leq \lambda_i}$.
Over the domain of $g_i^\star$, namely $|\alpha|\leq \lambda_i$, its subgradient is,
\begin{equation*}
	\partial g_i^*(\alpha) =
		\begin{cases}
			(-\infty, 0] & \alpha = -\lambda_i  \\
			0 & |\alpha| < \lambda_i \\
			[0, \infty) & \alpha = \lambda_i
		\end{cases}
    ~~.
\end{equation*}

We next show that $\forall i\in[n]$, there exists a mapping
$\ix\mapsto\ia_i(\ix)$, where $\ia_i(\cdot)$ is a continuous non-decreasing
piecewise linear function in $\ix$. This property serves as a succinct
representation of $S_i = \{(\ix, \ia_i(\ix))\}$ which takes the
following recursive form,
\beq{aix}
\ia_i(\ix) =
  \Big[\ia_{i+1}(\ix) + \ix - y_{i+1}\Big]_{-\lambda_i}^{+\lambda_i} ~~.
\eeq
To start, let us define $a_n(\ix) \equiv 0$ which adheres with the definition
of $S_n$ from the previous section. We next show that the form of $S_i$ holds
inductively, using the definitions of $S_i$ and $S_{i+\ha}$ from the previous
section. Assume that \eqr{aix} holds true for $S_{i+1}$. The definition of
the recursive map for $i+\ha$ implies that,
$$S_{i+\ha} = \{(\ix, \ia_{i+1}(\ix) + \ix - y_{i+1}) : \ix\in \R\}~~.$$
For convenience we define $\ia_{i+\ha}(\ix) = \ia_{i+1}(\ix) + \ix - y_{i+1}$.
In order to evaluate,
$$S_i = \{ (\ix, \ia) : \ix - \ix' \in \partial g_i^*(\ia)\;;\;
  \ia = \ia_{i+\ha}(\ix') \}$$
we need to analyze $\Delta\ix=\ix-\ix'$ for the admissible set of values of
$\ia$ under the map $g_i^*$ which is $[-\lambda_i, \lambda_i]$. Since
$\ia_{i+1}(\cdot)$ is continuous and non-decreasing, it implies that
$\ia_{i+\ha}(.)$ is strictly-increasing and bijective. Therefore, there exists two
unique boundary values $z_i^-\st\ia_{i+\ha}(z_i^-)=-\lambda_i$ and
$z_i^+\st\ia_{i+\ha}(z_i^+)=+\lambda_i$. For $\ia=\ia_{i+\ha}(\ix')$ such that
$-\lambda_i<\ia<\lambda_i$ we have $\partial g^*_i(\ia) = \{0\}$.
Thus, $\ix=\ix'$ and both are in $(z_i^-,z_i^+)$. We therefore get that
the restriction of $S_i$ to $\ix\in(z_i^-,z_i^+)$ yields,
\begin{eqnarray*}
S_i \cap \{\ix: z_i^- < \ix < z_i^+\}
  &=& \{(\ix, \ia_{i+\ha}(\ix)) : z_i^- < \ix < z_i^+\} \\
  &=& \{(\ix, \ia_{i+1}(\ix) + \ix - y_{i+1}): z_i^- < \ix < z_i^+\}
~~.
\end{eqnarray*}
We last need to address the cases $\ia=\pm \lambda_i$.
For $\ia=\ia_{i+\ha}(z_i^+)=\lambda_i$ we have $\partial g^*_i(\ia) = [0,\infty)$.
Thus, for any $\ix\geq z_i^+$ we have $\ix-z_i^+\in\partial g^*_i(\lambda_i)$
and therefore $(\ix,\lambda_i)\in S_i$. Analogously, for $\ia=-\lambda_i$ which
implies that for $\ix\leq z_i^-$, it holds that
$\ix-z_i^-\in\partial g^*_i(-\lambda_i) \Rightarrow (\ix,-\lambda_i)\in S_i$.
We therefore get that the restriction of $S_i$ to $\ix\geq z_i^+$ and
$\ix\leq z_i^-$ yields,
\begin{eqnarray*}
S_i \cap \{\ix: \ix\geq z_i^+\}
  &=& \{(\ix, \lambda_i) : \ix\geq z_i^+\} \\
S_i \cap \{\ix: \ix\leq z_i^-\}
  &=& \{(\ix, -\lambda_i) : \ix\leq z_i^-\}
~~.
\end{eqnarray*}
Since $\ia_{i+\ha}(\cdot)$ is increasing, we have for
$\ix \geq z_i^+ \Rightarrow \ia_{i+\ha}(\ix) \geq \lambda_i$ and
$\ix \leq z_i^- \Rightarrow \ia_{i+\ha}(\ix) \leq-\lambda_i$.
Therefore, we established that
$$ S_i =
  \left\{
    \Big(
      \ix,
      \Big[\ia_{i+1}(\ix) + \ix - y_{i+1}\Big]_{-\lambda_i}^{+\lambda_i}
    \Big) : \ix\in\R\right\} ~~.
$$
\begin{figure}[t]
\centerline{\fbox{\includegraphics[width=0.85\textwidth]{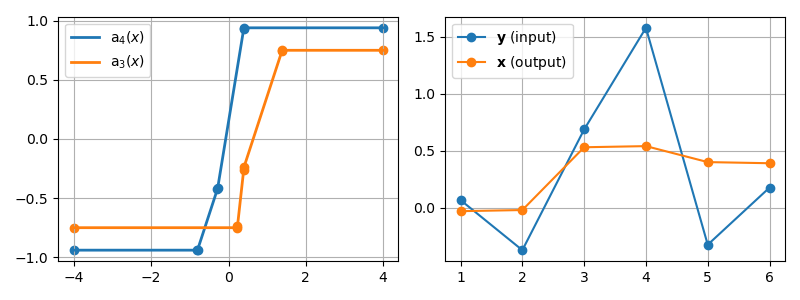}}}
\caption{Construction of $\ia_3(\cdot)$ from $\ia_4(\cdot)$
and optimal solution for the fused lasso.}
\end{figure}

\subsection{Sparse Fused Lasso}
We next combine both the standard lasso and fusion penalties.
Namely, we choose
$$ 
  h_i(x) = \frac{1}{2} (x - y_i)^2 + \beta_i |x| ~\mbox{ and }~
  g_i(\delta) = \lambda_i |\delta| ~~.
$$
These choices imply that
\begin{equation*}
	\partial h_i(x) =
		\begin{cases}
			x - y_i + \beta_i & x > 0  \\
			[-y_i - \beta_i, -y_i + \beta_i] & x = 0 \\
			x - y_i - \beta_i & x < 0
		\end{cases}
    ~~.
\end{equation*}
Generalizing the fused lasso, the succinct representation
for the sparse fused lasso is of the form,
$$S_i = \{(\ix,\ia_i^+(\ix)) : \ix > 0\}
  \,\cup\, \{(0, \ia) : \ia_i^-(0)\leq \ia\leq \ia_i^+(0)\}
  \,\cup\, \{(\ix,\ia_i^-(\ix)) : \ix < 0\}
~~.$$
The locally-admissible set consists of three disjoint subsets
induced by the sign of the first argument $\ix$. It is immediate
to verify that the subgradient at any value but zero is unique. Hence,
the mappings for positive and negative numbers are well-defined
and take similar recursive forms to that of the fused lasso,
$$
  \ia_i^\pm(\ix) =
  \Big[
    \ia_{i+1}^\pm(\ix) + \ix - y_{i+1} \pm \beta_{i+1}
  \Big]_{-\lambda_i}^{+\lambda_i} ~~.
$$
The sole value that requires further attention is at $\ix=0$.  The admissible
interval at zero is determined by the two boundary values, $\ia_i^-(0)$ and
$\ia_i^+(0)$. To facilitate an efficient implementation, we store the value of
$\Delta_i = \ia_i^+(0) - \ia_i^-(0)$. Prior to rectification,
we update $\Delta_i\gets \Delta_{i-1} + 2\beta_i$.  Analogous to the fused
lasso, we traverse the segments of $\ia_i(\cdot)$ left-to-right until we reach
the segment encapsulating $-\lambda_i$.  The only difference is that we
increase the value of $\ia_i(\cdot)$ by $\Delta_i$ upon crossing the zero.
Analogously, we decrease the value of $\ia_i(\cdot)$ by $\Delta_i$ upon
crossing the zero when traversing right-to-left.  After obtaining the critical
points $z_i^\pm$, we update $\Delta_i$ accordingly.  In each iteration, we
incur constant additional costs and thus the total runtime remains intact as
$\oo(n)$.

\begin{figure}[t]
\centerline{\fbox{\includegraphics[width=0.85\textwidth]{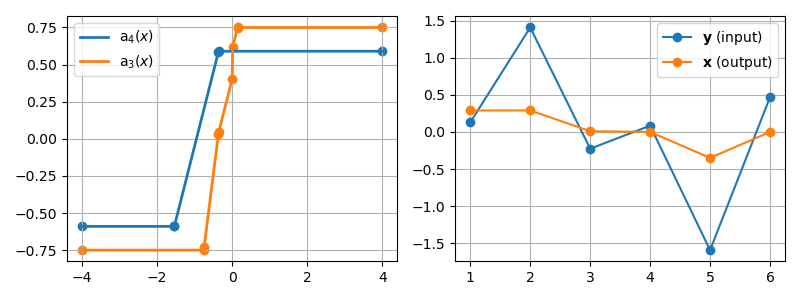}}}
\caption{Construction of $\ia_3(\cdot)$ from $\ia_4(\cdot)$
and optimal solution for the sparse fused lasso.}
\end{figure}

\subsection{Bregman Divergences} \label{bregdiv}
Before proceeding with nascent total variation regularizers, we now describe
a simple way for incorporating convex losses for $h_i(\cdot)$ at {\em zero}
additional computational cost. To remind the reader, given a strictly convex
function $\phi: \reals \to \reals$, the Bregman divergence between
$x\in\reals$ and $y\in\reals$ is defined as,
$D_\phi(x,y) = \phi(x) - (\phi(y) + \nabla\phi(y) (x - y))$. Our generalization
amount to setting, $h_i(x) = D_\phi(x, y_i)$ and keeping $g_i(\delta)$
intact, thus retaining for the fused lasso $g_i(\delta) = \lambda_i |\delta|$.
This choice for $h_i$ implies that
$\nabla h_i(x) = \nabla\phi(x) - \nabla\phi(y_i)$.
Since $\phi(\cdot)$ is strictly convex, $\nabla\phi(\cdot)$ is continuous and strictly
increasing. Following the same line of derivation as above we get that
the recursive mappings for each $\ia_i(\cdot)$  takes the following form,
\beq{aix_link}
\ia_i(\ix) =
  \Big[ \ia_{i+1}(\ix) + \nabla\phi(\ix) -
  \nabla\phi(y_{i+1}) \Big]_{-\lambda_i}^{+\lambda_i} ~~ .
\eeq
This implies that $\ia_i(.)$ is a piecewise linear function of $\nabla\phi(.)$.
We thus follow the same procedure above whilst replacing
$\ix\mapsto\nabla\phi(x)$ for each linear segment in $\ia_i$. This
generalization using a Bregman divergence is applicable throughout the
remainder of the section. We keep using the squared error for brevity of our
derivations.

\subsection{Isotonic Regression} \label{monoreg}
In isotonic regression the goal is to find the best non-decreasing sequence,
$x_1 \leq x_2 \leq \ldots \leq x_i \leq x_{i+1} \leq \ldots \leq x_n$ for which
the sum of the losses $h_i(x) = \frac{1}{2} (x - y_i)^2$ is minimized. We
replace the fused penalties with $g_i(\delta) = \1_{\delta\leq 0}$.
This choice implies that $g_i^*(\alpha) = \1_{\alpha\geq 0}$ and its
subgradient is,
\begin{equation*}
	\partial g_i^*(\alpha) =
		\begin{cases}
			(-\infty, 0] & \alpha = 0 \\
			0 & \alpha > 0
		\end{cases}
  ~~.
\end{equation*}
Parallel to the fused lasso, there exists a succinct representation for
isotonic regression of the form $S_i = \{(\ix, \ia_i(\ix)) : \ix\in \R\}$ which
is prescribed recursively as,
$$\ia_i(\ix)
  = \big[ \ia_{i+1}(\ix) + \ix - y_{i+1} \big]_+ $$
where $[z]_+ = \max(z, 0)$. The derivation follows the same lines as above
as there exists a unique boundary value $z_i^-\st\ia_{i+\ha}(z_i^-)=0$. The
resulting description of $S_i$ is obtained from the union of,
\begin{eqnarray*}
S_i \cap \{\ix: \ix > z_i^-\}
  &=& \{(\ix, \alpha_i'(\ix)) : \ix > z_i^-\} \\
S_i \cap \{\ix: \ix \leq z_i^-\}
  &=& \{(\ix, 0) : \ix \leq z_i^-\}
~~.
\end{eqnarray*}

We can in fact entertain a simple generalization which encompasses both
absolute difference penalties, isotonic regression, and asymmetric
difference penalties by defining,
$$g_i(\delta) = \bar{\lambda}_i [\delta]_+ \,-\, \underline{\lambda}_i [-\delta]_+
 \,\mbox{ with }\, \bar{\lambda}_i \geq 0 \,,\,  \underline{\lambda}_i \leq 0~~.$$
The recursive mapping takes the following form,
$$\ia_i(\ix) =
  \big[ \ia_{i+1}(\ix) + \ix - y_{i+1} \big]_{\underline{\lambda}_i}^{\bar{\lambda}_i}
~~.$$
Setting $-\underline{\lambda}_i=\bar{\lambda}_i=\lambda_i$ yields the fused
lasso whereas choosing $\underline{\lambda}_i=0$ and $\bar{\lambda}_i=\infty$
provides a template for isotonic regression. In addition, choosing
$\underline{\lambda}_i=0$ and $\bar{\lambda}_i=\lambda$ recovers the setting of
\cite{tibshirani2011nearly}. To conclude this section we would like to note
that isotonic regression can be used as the core for finding the closest vector
to $\mathbf{y}$ subject to a $\leq k$-modality constraint. Due to
submodularity of the objective the total runtime would be $\oo(kn)$.
\begin{figure}[t]
\centerline{\fbox{\includegraphics[width=0.85\textwidth]{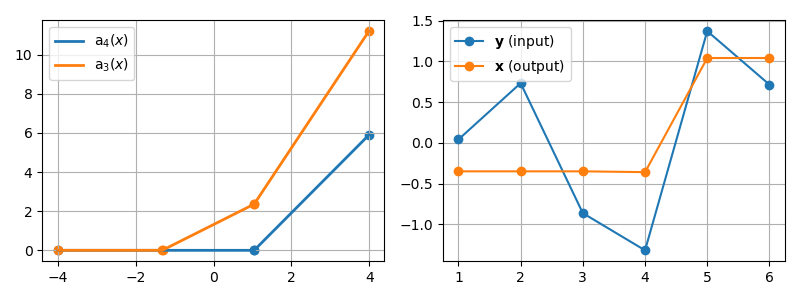}}}
\caption{Construction of $\ia_3(\cdot)$ from $\ia_4(\cdot)$
and optimal solution for isotonic regression.}
\end{figure}

\subsection{Fused Barriers} \label{barrier}
For certain data analysis applications the goal is to eliminate outlier points
in an observed sequence rather than smooth or filter the sequence. We cast
the problem as requiring the differences between consecutive points not to
exceed prescribed maximal changes $b_i$ by establishing a series of barrier
functions, $g_i(\delta) = \1_{|\delta|\leq \lambda_i}$. Each $b_i$ here can be
thought of as the maximum tolerance to changes and is not necessarily small and
thus restrictive. For this choice of $g_i$ we get,
\begin{equation}
	\partial g_i^*(\alpha) =
		\begin{cases}
      -\lambda_i       & \alpha < 0 \\
      [-\lambda_i,\lambda_i] & \alpha = 0 \\
			\lambda_i        & \alpha > 0
		\end{cases} ~~. \eqd{gstar_barrier}
\end{equation}

Albeit being slightly more involved, we next show that there exists a succinct
representation of $S_i$ in the form of a piecewise linear map, $\ix \mapsto
\ia_i(\ix)$.  First, let us reuse the intermediate function $\ia_{i+\ha}$ from the
derivation of the fused lasso algorithm, $\ia_{i+\ha}(\ix) = \ia_{i+1}(\ix)+\ix
-y_{i+1}$. We denote by $\kappa_\lambda(\cdot)$ the \kvn\ operator (from
Yiddish, literally “to squeeze, shrink”), colloquially called shrinkage,
$$\kappa_\lambda(z) = \sign(z) \, \big[|z| - \lambda\big]_+ ~~.$$
The construction of $\ia_i$ from $\ia_{i+1}$ consists of the following steps,
\begin{align}
  \ia_{i+\ha}(\ix) &:=  \ia_{i+1}(\ix) + \ix - y_{i+1}  \tag{Add Coordinate} \\
    z_i       &:=  z \, \st \ia_{i+\ha}(z) = 0  \tag{Zero Crossing} \\
   \ia_i(\ix) &:= \ia_{i+\ha}\big( z_i + \kappa_{\lambda_i}(\ix - z_i) \big)
   \tag{\kvn} \label{kvn}
\end{align}
Informally, after constructing the intermediate function $\ia'_i$, which is
strictly monotone, we find the point $z_i$ where $\ia'_i$ crosses zero. We
then insert an interval of zero slope whose center is $z_i$ and its support
spans from $z_i-\lambda_i$ through $z_i+\lambda_i$.

We now provide the details of the derivation of $\ia_i$. Since $z_i$ is the
locus of the zero of $\ia'_i$ then for $\ix'>z_i$ we have
$\ia=\ia_{i+\ha}(\ix')>0 \Rightarrow \partial g_i^*(\ia)=\lambda_i$ based on
\eqr{gstar_barrier}.
Thus, for any $\ix=\ix'+\lambda_i>z_i+\lambda_i$ we have $(\ix,\ia_{i+\ha}(\ix-\lambda_i))\in S_i$.
Analogously, for $\ix<z_i-\lambda_i$ we have $(\ix,\ia_{i+\ha}(\ix+\lambda_i))\in S_i$.
Last, for $(z_i,\ia'_i(z_i))=(z_i,0)$ we have $\partial g_i^*(0)=[-\lambda_i,\lambda_i]$.
Therefore, for any $\ix$ such that $z_i-\lambda_i\leq \ix\leq z_i+\lambda_i$ it holds that
$\ix-z_i\in \partial g_i^*(0) \Rightarrow (\ix,0)\in S_i$.
The resulting $S_i$ is obtained from the union of,
\begin{eqnarray*}
S_i \cap \{\ix\,:\, \ix < z_i-\lambda_i\}
  &=& \{(\ix,\,\ia_{i+\ha}(\ix-\lambda_i)) \,:\, \ix < z_i-\lambda_i\} \\
S_i \cap \{\ix\,:\, z_i-\lambda_i \leq \ix \leq z_i+\lambda_i\}
  &=& \{(\ix,\,0) \,:\, z_i-\lambda_i \leq \ix \leq z_i+\lambda_i\} \\
S_i \cap \{\ix\,:\, \ix > z_i+\lambda_i\}
  &=& \{(\ix,\,\ia_{i+\ha}(\ix+\lambda_i)) \,:\, \ix > z_i+\lambda_i\}
~~.
\end{eqnarray*}
Eq.~({\kvn}) provides a succinct functional form of the union of these three
sets.

Since each $\ia_i(\cdot)$ is a piecewise linear function we can employ the same
algorithmic skeleton used for the fused lasso. Alas, in order to construct
$\ia_i$ we need to traverse $\ia'_i$ and find the locus, $z_i$, of its zero.
This search would require $\oo(n)$ time and thus the total run time would
amount to $\oo(n^2)$. We provide in Appendix~\ref{skipl:app} a description of a
more efficient implementation which uses a more elaborated data structure in
order to reduce the amortized runtime to $\oo(n\log(n))$.
\begin{figure}[t]
\centerline{\fbox{\includegraphics[width=0.85\textwidth]{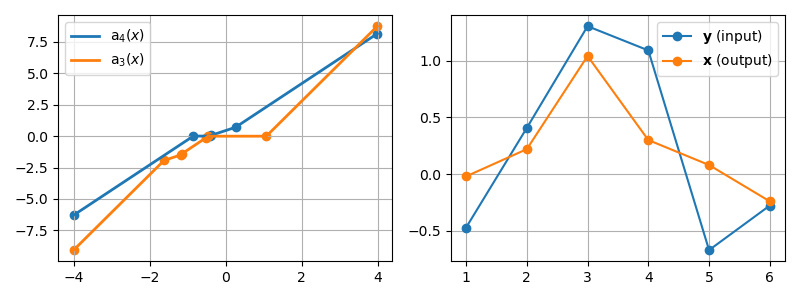}}}
\caption{Construction of $\ia_3(\cdot)$ from $\ia_4(\cdot)$ and optimal
solution for fused barriers.} \end{figure}

We conclude the section with a short discussion and illustration of the
characteristics of the variational penalties of the fused lasso by contrasting
it with barrier constraints on the variation. To this end, we generated a
sequence based on a quadratic function, $y_i = (i - 50)^2 / 100$. We then
contaminated the sequence with noise. We first added i.i.d noise to each
element sampled from the uniform distribution over $[-\epsilon,\epsilon]$. We
next added ``shock'' noise by replacing the value of a few elements chosen
at random with a small {\em negative number}. The generating and noisy
sequences are plotted on the left hand-side of Figure~5. We then solved least
squares approximation for a range of variational penalties and constraints by
varying $\lambda$. For each solution $\x(\lambda)$ we calculated the $\ell_1$
and $\ell_\infty$ variation norm. The results are shown on the right hand-side
of Figure~5. The dashed two-sided arrows designate solutions attaining the same
approximation error by the two methods. It is evident that for the same error
the fused lasso naturally obtains a lower $\ell_1$ variation norm
and it is more resilient to the uniform noise. In contrast, posing
a barrier constraint on the variation provides a uniformly lower
$\ell_\infty$ norm of the variation and is more resilient to shock noise.
\begin{figure}[t]
\centerline{\fbox{\includegraphics[width=0.95\textwidth]{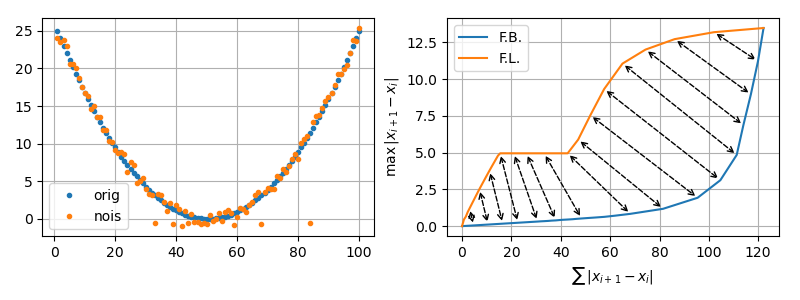}}}
\caption{Comparison least-squares sequence approximation with variational
penalty of the fused lasso (FL) versus barrier constraints (FB).}
\end{figure}

\section{High-Order Variational Penalties} \label{hiord:sec}
The differences $x_{i+1}-x_{i}$ can morally be viewed as
\href{https://en.wikipedia.org/wiki/Discrete_calculus}{discrete
derivatives}. The focus of this section is higher order discrete
derivatives as variational penalties. For example, penalizing
for the second order derivative amounts to,
\begin{equation*}
\frac12\|\x-\y\|^2
+ \sum_{i=1}^{n-2} \lambda_i \big|(x_{i+2} - x_{i+1}) - (x_{i+1} - x_i)\big| ~~.
\end{equation*}
Rather than discussing specific forms of discrete difference operators, we
analyze a general case stemming from convolution operators. Formally, let
$\bl=(-1, \ll)$ where $\ll = (\convf_1, \convf_2, \ldots, \convf_k)$ be
vectors over $\reals^{k+1}$ and $\reals^k$ respectively. The convolution of
$\x=(x_1, \ldots, \x_n)$ with $\bl$ is a vector whose $i$'th entry is,
$(\bl \ast \x)_{i} = \sum_{m=0}^k \convf_m\,x_{i-m}$.

Equipped with these definitions we now focus
on problems of the following general form,
\beq{convopt}
  \min_\x \sum_{i=1}^n h_i(x_i) +
    \sum_{i=k}^{n-1} g_i\big((\bl \ast \x)_{i+1}\big) ~~.
\eeq

Since scaling is a degree of freedom we set $\convf_0=-1$ and use the
definition above $\bl=(-1,\ll)$ to get that $(\bl\ast\x)_i=(\ll\ast\x)_i-x_i$.
Here we on purpose shift the indices of $g_i$ by $1$ to keep the same notation
as used in the algorithms above when plugging in $\ll=(1)$. We use
$\bl\circledast\v$ to denote the convolution of a vector $\v$ with the reversal
of $\bl$ which we define as,
$$(\bl\circledast\v)_i = \sum_{m=0}^k \convf_m v_{i+m} ~~.$$

We again confine ourselves to settings for which strong duality holds thus,
\begin{eqnarray*}
U & = & \min_{\x_{1:k}}\, \max_{\alpha_k}\, \min_{x_{k+1}}
	\ldots \max_{\alpha_{n-1}}\, \min_{x_n}\: \sum_{i=1}^n h_i(x_i)
	+ \sum_{i=k}^{n-1} \left[ \alpha_i (\bl* \x)_{i+1} - g_i^*(\alpha_i) \right]
\end{eqnarray*}
Analogously, we define a sequence of intermediate functions,
\begin{eqnarray} \label{fixi2}
f_i(\x) & \equiv & \max_{\alpha_i}\,
	\ldots \min_{x_n}\: \sum_{j=i}^{n-1}
	\left[ h_{j+1}(x_{j+1}) + \alpha_j (\bl* \x)_{j+1} - g_j^*(\alpha_j)
	\right] ~~.
\end{eqnarray}
This definition means that $f_i:\reals^i\to\reals$ and thus formally defined
as $f_i(\x_{1:i})$ since $\x_{i+1:n}$ is grounded to its optimal vector.
As boundary conditions we set $f_n(\x) = 0$. Similar to the constructions
above we introduce an auxiliary function, $f_{i-\ha}(\x) = f_i(\x) + h_i(x_i)$
which, like $f_i$, is a function of $\x_{1:i}$.

To simplify notation we denote by $\ixb_i$ the prefix $\x_{1:i}$
and $\iab_i$ for $\ba_{i:n-1}$. Albeit the analogous notation,
$\ixb_i$ denotes a free variable whereas $\iab_i$ designates the {\em
optimal} dual vector.
Using these definitions we introduce a summation-free form for $f_i$,
\begin{eqnarray}
f_i(\x) & = & \max_{\alpha_i}\:
  \alpha_i (\ll* \x)_{i+1} - g_i^*(\alpha_i)
  + \min_{x_{i+1}} \left\{f_{i+1}(\x) + h_{i+1}(x_{i+1})
  - \alpha_i x_{i+1}  \right\} \nonumber \\
& = & \max_{\alpha_i}\:
  \alpha_i (\ll* \x)_{i+1} - g_i^*(\alpha_i)
  + \min_{x_{i+1}} \left\{f_{i+\ha}(x_{i+1}|\ixb_i)
  - \alpha_i x_{i+1} \right\} \nonumber \\
& = & \max_{\alpha_i}\:
  \alpha_i (\ll* \x)_{i+1} - g_i^*(\alpha_i)
	- f_{i+\ha}^*(\alpha_i|\ixb_i) ~~,
\end{eqnarray}
where $f_{i+\ha}(\cdot|\ixb_i)=f_{i+\ha}(\ixb_i,\cdot)$.

The subgradient conditions for locally-restricted optimality define
two admissible sets,
\begin{eqnarray*}
S_{i+\ha} & = &
  \big\{ (\ixb_{i+1}, \iab_i) : \ia_i \in \partial
  f_{i+\ha}(\ix_{i+1}|\ixb_i) \big\} \\
S_i & = & \big\{ (\ixb_i, \iab_i)
  : (\ll* \ixb)_{i+1} \in \partial f_{i+\ha}^*(\ia_i|\ixb_i)
  + \partial g_i^*(\ia_i) \big\} ~~. \\
\end{eqnarray*}
We next show that the following recursive form holds,
\begin{eqnarray*}
S_{i+\ha} & = &
  \big\{ (\ixb_{i+1}, \iab_i) : \ia_i
  - (\ll\circledast \iab)_i
  \in \partial h_{i+1}(\ix_{i+1})
    \;;\; (\ixb_{i+1}, \iab_{i+1}) \in S_{i+1} \big\} \\
S_i & = & \big\{ (\ixb_i, \iab_i)
  : (\ll* \ixb)_{i+1} - \ix_{i+1} \in \partial g_i^*(\ia_i)
    \;;\; (\ixb_{i+1}, \iab_i) \in S_{i+\ha} \big\} ~~.
\end{eqnarray*}
Note that the admissible set consists of the full sequence of
locally-admissible pairs. From strong duality, $S_i$ de-facto defines a
point-to-set mapping from a prefix vector $\ixb_i$ to its dual solutions of
$f_i(\cdot)$. Hence, we can rewrite~Eq.~\eqref{fixi2} as,
\begin{eqnarray*}
f_i(\x) & = & \max_{\ba_{i:n-1}}\,
  \min_{\x_{i+1:n}}\: \sum_{j=i}^{n-1}
  \left[ h_{j+1}(x_{j+1}) + \alpha_j (\bl* \x)_{j+1} - g_j^*(\alpha_j)
  \right] \\
& = & \max_{\ba_{i:n-1}}\: \sum_{j=i}^{n-1} \left[
  \alpha_j (\bl* \ixb_i)_{j+1}
  - h_{j+1}^*(-(\bl\circledast \ba)_j)
  - g_j^*(\alpha_j) \right] ~~.
\end{eqnarray*}
Thus, given $\ixb_i$, for any $\iab_i$ attaining the above maximum we have
$$\forall m\in \{0,\ldots,k-1\}: \;
  (\ll\circledast \iab_i)_{-m} \in \partial_{x_{i-m}} f_i(\ixb_i) ~ .$$
Now, given $S_{i+1}$ we perform the following expansion,
\begin{eqnarray*}
S_{i+\ha} & = &
  \big\{ (\ixb_{i+1}, \iab_i) : \ia_i \in \partial
  f_{i+\ha}(\ix_{i+1}|\ixb_i) \big\} \\
& = & \big\{ (\ixb_{i+1}, \iab_i) : \ia_i \in \partial
  f_{i+1}(\ix_{i+1}|\ixb_i) + \partial h_{i+1}(x_{i+1}) \big\} \\
& = & \big\{ (\ixb_{i+1}, \iab_i) :
  \ia_i - (\ll\circledast \iab)_i
  \in \partial h_{i+1}(x_{i+1})
  \;;\; (\ixb_{i+1}, \iab_{i+1}) \in S_{i+1} \big\} ~~.
\end{eqnarray*}
Using the fact that $\ia_i\in\partial f_{i+\ha}(\ix_{i+1}|\ixb_i)
~\Leftrightarrow~ \ix_{i+1}\in\partial f^*_{i+\ha}(\ia_i|\ixb_i)$
we get,
\begin{eqnarray*}
S_i & = &
  \big\{ (\ixb_i, \iab_i)
  : (\ll* \ixb)_{i+1} \in \partial f_{i+\ha}^*(\ia_i|\ixb_i)
  + \partial g_i^*(\ia_i) \big\} \\
& = & \big\{ (\ixb_i, \iab_i)
  : (\ll* \ixb)_{i+1} - \ix_{i+1} \in \partial g_i^*(\ia_i)
    \;;\; (\ixb_{i+1}, \iab_i) \in S_{i+\ha} \big\} ~~.
\end{eqnarray*}
Finally, to satisfy the boundary conditions we need to find the solution of,
$$\min_{\x_{1:k}} f_k(\x) + \sum_{i=1}^k h_i(x_i) ~~.$$
The subgradient conditions imply that
$$\forall i\in \{1,\ldots, k\}:\,
0\,\in\, (\ll\circledast \iab_k)_{i-k} + \partial h_i(\ix_i)$$
where we used the above property to compute $\partial_{\ix_i} f_k(\x)$.

\begin{figure}[t!]
\centering
\fbox{
\begin{subfigure}{.48\textwidth}
  \centering
  \includegraphics[width=\linewidth]{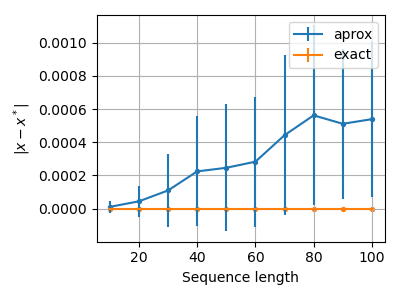}
\end{subfigure}%
\begin{subfigure}{.48\textwidth}
  \centering
  \includegraphics[width=\linewidth]{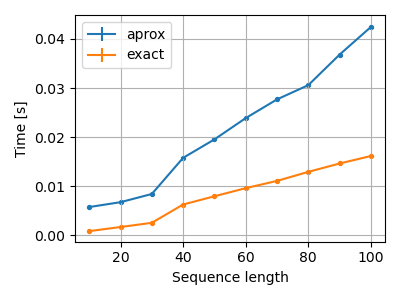}
\end{subfigure}
}
\caption{Exact vs. approximate subgradient following for
second-order variational penalties.}
\end{figure}
\paragraph{Second-order Variational Penalties.} To illustrate the power of
the convolution-based representation, we provide a specific derivation for
variational penalties which correspond to second order finite differences.
To do so we simply need to set $\ll = (2, -1)$. Since $f_i(\cdot)$ depends only on
$x_{i-1}$ and $x_i$, we can represent the subgradient sets as,
$$S_i = \{\ia_i(\ix_{i-1}, \ix_i)\;;\; \ia_{i+1}(\ix_{i-1}, \ix_i) \}
~\mbox{ and }~
S_{i+\ha} = \{\ia_{i+\ha}(\ix_i, \ix_{i+1})\;;\; \ia_{i+\frac32}(\ix_i, \ix_{i+1}) \}~~.$$
Assume this holds true for $S_{i+1} ~\mbox{ i.e. }~
S_{i+1} = \{\ia_{i+1}(\ix_i, \ix_{i+1})\;;\; \ia_{i+2}(\ix_i, \ix_{i+1}) \}$.
The definition of the recursive map for $i+\ha$ implies that,
$$\ia_{i+\ha}(\ix_i,\ix_{i+1}) = 2\ia_{i+1}(\ix_i,\ix_{i+1})
- \ia_{i+2}(\ix_i,\ix_{i+1}) + \ix_{i+1} - y_{i+1}
~\mbox{ and }~
\ia_{i+\frac32}(\cdot) = \ia_{i+1}(\cdot)~~.$$
To obtain the recursive map $S_{i+\ha}\to S_i$, we need to extract
$\ix_{i+1}$ from the following subgradient set equation,
$$2 \ix_i - \ix_{i-1} - \ix_{i+1}
  \in \partial g_i^*(\ia_{i+\ha}(\ix_i, \ix_{i+1}))~~.$$
Similar to the fused lasso, the solution is given by
the backward map $T_{i+1}(\ix_{i-1},\ix_i)$ defined as
$$ T_{i+1}(\ix_{i-1},\ix_i) =
  \Big[2\ix_i - \ix_{i-1} \Big]_{z_{i+1}^-(\ix_i)}^{z_{i+1}^+(\ix_i)}
~~, $$
where $z_{i+1}^\pm(\ix_i)$ denotes the boundary values such
that $\ia_{i+\ha}(\ix_i, z_{i+1}^\pm(\ix_i)) = \pm 1$.
Using this form of $T$ back in $\ix_{i+1}$, we obtain
$$\ia_i(\ix_{i-1},\ix_i) = \ia_{i+\ha}(\ix_i, T_{i+1}(\ix_{i-1},\ix_i))$$
and
$$\ia_{i+1}(\ix_{i-1},\ix_i) = \ia_{i+\frac32}(\ix_i, T_{i+1}(\ix_{i-1},\ix_i))~~.$$

Here, $\ia_i(\cdot)$ is a piecewise linear manifold in $\reals^2$.
We store triplets of the form $\{v_j^1, v_j^2, \alpha_j\}$ for
vertices of each facet defining $\ia_i$.
We conjecture that the number of facets grows as $\oo(n^2)$ but
leave formal analysis of runtime as future research.
We empirically observed that the description of $\ia_i$ is typically
linear in the length of the input sequence. Alas, a disadvantage
of the exact subgradient following described here is the need to maintain
a data structure for high dimensional objects.
We describe an alternative approximation yielding a simple iterative
procedure for the problem in Appendix~\ref{high_order:app}.

We next demonstrate the effectiveness of exact subgradient following over the
approximation-based method. To do so, we generated synthetic data of varying
sequence length similar to the previous section. We sampled
$(x_{i+2}-x_{i+1})-(x_{i+1}-x_i)$ from $\mathcal{D}$ and then constructed $\y$
using the optimality conditions. Throughout the experiment, the sparsity level
was set to $50\%$. In each experiment, we ran the approximation-based algorithm
for $100$ iterations. The results are summarized in Figure~8. The advantage of
exact subgradient following over approximation both in terms of runtime and
optimality is apparent. The exact subgradient following weakly depends on the
problem size whereas the quality of the approximate solution as well as run
time quickly deteriorates even for short sequences.

\section{Variationally Penalized Multivariate Regression} \label{sec:mv}
We next extend the subgradient following framework to multivariate
settings. For concreteness we examine objectives for instance $i$
of the form,
\beq{Qxy}
h_i(\x)
  \,=\, \frac{1}{2} \big\| \x - \y_i \big\|^2_{{Q}_i}
  \,=\, \frac{1}{2} (\x - \y_i)^\top \m{Q}_i (\x - \y_i)
  ~ ~\x,\y_i\in\reals^n~~.
\eeq
We assume w.l.o.g that the quadratic form is strictly positive definite,
$\m{Q}_i\succ 0$. The multivariate setting is intrinsically more complex than
the scalar case due to the penalty entanglement of $\x_i$ with itself and with
$\x_{i+1}$. To mitigate this difficulty we first examine the squared Euclidean
distance between $\x_{i}$ and $\x_{i+1}$,
$$g_i(\bd)
  \,=\, \frac{1}{2} \lVert  \x_{i+1} - \x_i \rVert^2
  \,=\, \frac{1}{2} \lVert \bd\rVert^2 ~~.$$
This setting generalizes the squared fused penalty of the scalar case for which
$Q_i = [q_i\in\reals]$ and thus can be absorbed into $\lambda_i$. With these
assumptions we get,
$$\nabla h_i(\x) =
  \m{Q}_i (\x - \y_i) ~\mbox{and}~ \nabla g_i^*(\bd) = \bd ~~.$$
The derivation of the subgradient following for this setting would also serve
us as a building block in more complex multivariate problems described and
analyzed in the sequel.

The Fenchel dual for the multivariate case where $f:\reals^n\to\reals$ is
defined as,
$$
  f^*(\ba) =
    \sup_{\x\in\mathrm{dom}(f)} \big\{\ba^\top \x - f(\x)\big\} ~~.
$$
Both $h_i(\cdot)$ and $g_i^*(\cdot)$ defined above are strictly convex,
as we require $\m{Q}_i\succ 0$, thus the Fenchel duals for all $i\in[n]$ have
a unique maximizer that forms dual feasible pairs $(\x,\ba)$. The equation
$\ba=\nabla\!f(\x)$ is typically referred to as the link function or mirror map.
We use this property in the ensuing derivation.

Since the components of the gradients are of unary form and linear, the
subgradient sets can be written succinctly as,
$$S_i =
  \left\{ (\x, \iab) : \m{A}_i \x + \m{B}_i \iab + \c_i = \0 \right\} ~~.$$
Furthermore, as the mirror map here is bijective, the admissible sets of~Eq.~\eqref{Sih}
and~Eq.~\eqref{Si} can be written as,
\begin{eqnarray*}
S_{i+\ha} & = &
  \big\{ (\ixb_{i+1}, \iab_i) : \iab_i = \nabla\! f_{i+\ha}(\ixb_{i+1}) \big\} \\
S_i & = & \big\{ (\ixb_i, \iab_i)
	: \iab_i = \nabla\! f_i(\ixb_i) \big\} ~~.
\end{eqnarray*}
We therefore effortlessly unravel the recursive form for $\m{A}_{i-1}$,
$\m{B}_{i-1}$, $\c_{i-1}$ from $\m{A}_{i}$, $\m{B}_{i}$, $\c_{i}$,
\vspace{-6pt} \\
\begin{minipage}{0.8\textwidth}
\setlength\baselineskip{16pt}
\begin{eqnarray*}
S_{i-\ha} & = & \big\{ (\ixb, \iab)
  : \iab - \iab' = \m{Q}_i (\ixb - \y_i);\;
    (\ixb, \iab') \in S_i \big\} \\ \smallskip
 & = & \big\{ (\ixb, \iab)
  : \m{A}_i \ixb + \m{B}_i \left(\iab - \m{Q}_i (\ixb - \y_i) \right)
    + \c_i = \0 \big\} \\ \smallskip
 & = & \big\{ (\ixb, \iab)
  : (\m{A}_i - \m{B}_i \m{Q}_i) \ixb + \m{B}_i \iab
    + (\c_i + \m{B}_i \m{Q}_i \y_i) = \0 \big\} ~~, \\ \smallskip
S_{i-1} & = & \big\{ (\ixb, \iab)
  : \ixb - \ixb' = \iab;\; (\ixb', \iab) \in S_{i-\ha} \big\} \\ \smallskip
 & = & \big\{ (\ixb, \iab)
  : (\m{A}_i - \m{B}_i \m{Q}_i) (\ixb - \iab)
    + \m{B}_i \iab + (\c_i + \m{B}_i \m{Q}_i \y_i) = \0 \big\} \\ \smallskip
 & = & \big\{ (\ixb, \iab)
  : (\m{A}_i - \m{B}_i \m{Q}_i) \ixb
    + (\m{B}_i (\m{I} + \m{Q}_i) - \m{A}_i) \iab
    + (\c_i + \m{B}_i \m{Q}_i \y_i) = \0 \big\} ~~.
\end{eqnarray*}
\end{minipage} \\
In summary, we get that,
\begin{eqnarray*}
\m{A}_{i-1} & = & \m{A}_i - \m{B}_i \m{Q}_i \\
\m{B}_{i-1} & = & \m{B}_i (\m{I} + \m{Q}_i) - \m{A}_i \\
\c_{i-1} & = & \c_i + \m{B}_i \m{Q}_i \y_i ~~.
\end{eqnarray*}
For boundary conditions we require,
$$(\m{A}_n, \m{B}_n, \c_n) = (\m{O},\, \m{I},\, \0) ~\mbox{ and } ~
  (\m{A}_1 - \m{B}_1 \m{Q}_1) \x_1 + (\c_1 + \m{B}_1 \m{Q}_1 \y_1) = 0  ~~.$$
Upon obtaining $\x_1$ along with $\ba_0\equiv \0$, we construct the trajectory
of the solution using the backward recursion,
\begin{eqnarray*}
\ba_i &=& \ba_{i-1} - \m{Q}_i (\x_i - \y_i) \\
\x_{i+1} &=& \x_i - \ba_i ~~.
\end{eqnarray*}

\section{Sparse Variationally Penalized Multivariate Regression}
\label{fusedmv:sec}
We henceforth use $\x_{1:n}$ to denote the sequence of vectors
$\x_1,\ldots,\x_n$. While the algorithm of Sec.~\ref{sec:mv} entertains a
concise form, it does not yield variational sparsity. To do so we can readily
use the $1$-norm as the variational penalty. This would yield though
element-wise fusion whereas full variational sparsity entails that for some
indices $i\in[n]$, $\x_i = \x_{i+1}$. In this section we use the algorithm from
the previous section as a building block while replacing the norm-squared
penalty $\|\x_{i+1} - \x_i\|^2$ with either the $2$-norm $\|\x_{i+1} - \x_i\|$
of the difference vector or its infinity norm $\|\x_{i+1} - \x_i\|_\infty$.
For brevity, we assume throughout this section that $\forall i, \lambda_i=1$
and are thus omitted.

\subsection{Multivariate Regression with $\ell_2$-Variational Penalty}
When we replace each of the norm-squared variational penalties with the $2$-norm,
the goal is to find the minimizer of,
\begin{equation}
\phi(\x_{1:n}) =
\frac12 \sum_{i=1}^n \|\x_i-\y_i\|^2
+ \sum_{i=1}^{n-1} \| \x_{i+1} - \x_i\| ~~.  \label{multi_fused:eq}
\end{equation}
Alas, this formulation does {\em not} entail a closed form subgradient
following procedure. The approach that we take here is to construct
a relaxation that yields the following convergent algorithm.
\begin{center}
\begin{minipage}{.72\linewidth}
\begin{algorithm}[H]
\label{alg:surr-mv}
\caption{Surrogate for Multivariate \moss}
\begin{algorithmic}
\STATE{\bf initialize} $\x_{1:n}^{0} = \y_{1:n}$
\FOR{$t = 1\ \textbf{to}\ t_{\text{max}}$}
\STATE $\forall i: \rho_i^{t} =
  \max\big(\|\x_{i+1}^{t} - \x_i^{t}\|,\, \epsilon\big)$
\STATE $\x^{t+1}_{1:n} =
  \displaystyle \argmin_{\x_{1:n}}
\frac12 \sum_{i=1}^n \|\x_i-\y_i\|^2
+ \sum_{i=1}^{n-1} \frac{\|\x_{i+1} - \x_i\|^2}{2 \rho_i^{t}} $
\ENDFOR
\STATE \textbf{return} $\x^{t_{\text{max}}}_{1:n}$
\end{algorithmic}
\end{algorithm}
\end{minipage}
\end{center}

We next show that Algorithm~1 converges linearly to the minimum of,
\begin{equation}
\hat{\phi}(\x_{1:n}) =
\frac12 \sum_{i=1}^n \|\x_i-\y_i\|^2
+ \sum_{i=1}^{n-1} g_{\epsilon}(\x_{i+1} - \x_i) ~~.  \label{approx_fused:eq}
\end{equation}
where $g_\epsilon(\cdot)$ is the Huber~\cite{huber1992robust} loss defined as,
\begin{equation*}
	g_\epsilon(\bd) =
		\begin{cases}
			\frac{\|\bd\|^2}{2\epsilon}  & \|\bd\|\leq \epsilon \\
			\|\bd\|-\frac{\epsilon}{2} & \|\bd\| > \epsilon
		\end{cases}
    ~~.
\end{equation*}

Since for any $\bd$ it holds that
$\|\bd\|-\frac{\epsilon}{2}\leq g_\epsilon(\bd) \leq \|\bd\|$,
the solution of the relaxed problem~Eq.~(\ref{approx_fused:eq})
is at most $\frac{(n-1)\epsilon}{2}$ suboptimal with respect to
that of the original problem defined in Eq.~(\ref{multi_fused:eq}).
Let us introduce a further auxiliary function,
\begin{equation*}
 \hat{g}_\epsilon(\bd,\bd') =
 	\begin{cases}
   \frac{\|\bd\|^2}{2\epsilon}  & \|\bd'\|\leq \epsilon \\
 	 \frac{\|\bd\|^2}{2\|\bd'\|} + \frac{\|\bd'\|-\epsilon}{2} &
    \|\bd'\| > \epsilon
	\end{cases}
 ~~.
\end{equation*}
The function $\hat{g}_\epsilon$ is a quadratic upper bound
on $g_\epsilon(\bd)$ expanded at $\bd'$, thus
$\forall\bd,\bd': ~ \hat{g}_\epsilon(\bd,\bd')\geq g_\epsilon(\bd)$
and $\hat{g}_\epsilon(\bd,\bd) = g_\epsilon(\bd)$.
By examining each case we get the following,
\begin{eqnarray*}
\hat{g}_\epsilon(\bd,\bd') &=& g_\epsilon(\bd')
  + \nabla{g_\epsilon}(\bd')^\top (\bd-\bd')
  + \frac{1}{2\,\max\big(\|\bd'\|, \epsilon\big)} \|\bd-\bd'\|^2 \\
&\leq& g_\epsilon(\bd) +
  \frac{1}{2\,\max\big(\|\bd'\|, \epsilon\big)} \|\bd-\bd'\|^2 ~~.
\end{eqnarray*}

Hence, Algorithm~1 is a proximal method where we iteratively apply the
multivariate subgradient-following of the previous section to obtain an exact
solution for each adaptively constructed upper bound. We follow a similar
skeletal proof to show the convergence.

We denote by $\eta^{t} = \min_i \rho_i^{t} \geq \epsilon$ and
the optimum of the relaxed problem as
$\hat{\phi}^*=\hat{\phi}(\x_{1:n}^*)$. We derive a series of
bounds as the means to show convergence, starting with the
definitions of $g_\epsilon$ and $\hat{g}_\epsilon$,
\begin{eqnarray*}
\hat{\phi}(\x_{1:n}^{t+1})
&=& \frac12\sum_{i=1}^n \|\x_i^{t+1}-\y_i\|^2
+ \sum_{i=1}^{n-1} g_\epsilon(\x_{i+1}^{t+1} - \x_i^{t+1}) \\
&\leq& \frac12\sum_{i=1}^n \|\x_i^{t+1}-\y_i\|^2
+ \sum_{i=1}^{n-1} \hat{g}_\epsilon(\x_{i+1}^{t+1}-\x_i^{t+1}, \x_{i+1}^{t}-\x_i^{t}) \\
&=& \min_{\x_{1:n}} \; \frac12\sum_{i=1}^n \|\x_i-\y_i\|^2
+ \sum_{i=1}^{n-1} \hat{g}_\epsilon(\x_{i+1}-\x_i, \x_{i+1}^{t}-\x_i^{t}) \\
&\leq& \min_{\x_{1:n}} \; \frac12\sum_{i=1}^n \|\x_i-\y_i\|^2
+ \sum_{i=1}^{n-1} g_\epsilon(\x_{i+1}-\x_i)
+ \frac{1}{2\rho_i^{t}} \|(\x_{i+1}-\x_i) - (\x_{i+1}^{t}-\x_i^{t})\|^2 \\
&\leq& \min_{\x_{1:n}} \; \frac12\sum_{i=1}^n \|\x_i-\y_i\|^2
+ \sum_{i=1}^{n-1} g_\epsilon(\x_{i+1}-\x_i)
+ \frac{1}{\rho_i^{t}} \big(\|\x_{i+1}-\x_{i+1}^{t}\|^2 + \|\x_i-\x_i^{t}\|^2\big) \\
&\leq& \min_{\x_{1:n}} \; \frac12\sum_{i=1}^n \|\x_i-\y_i\|^2
+ \sum_{i=1}^{n-1} g_\epsilon(\x_{i+1}-\x_i)
+ \sum_{i=1}^n \frac{2}{\eta^{t}} \|\x_i-\x_i^{t}\|^2 ~~.
\end{eqnarray*}
Next we introduce a auxiliary sequence
$\z_{1:n}(\mu) = (1-\mu)\,\x_{1:n}^{t}+\mu\,\x_{1:n}^*$ and upper bound
the last inequality using a minimizer constrained to the line segment
from $\x_{1:n}^{t}$ to $\x_{1:n}^{*}$. We then use the convexity of the
objective. These two steps amount to,
\begin{eqnarray*}
\hat{\phi}(\x_{1:n}^{t+1})
&\leq& \min_{\mu\in[0,1]} \;
   \frac12\sum_{i=1}^n \|\z_i(\mu)-\y_i\|^2
  + \sum_{i=1}^{n-1} g_\epsilon(\z_{i+1}(\mu)-\z_i(\mu))
  + \sum_{i=1}^n \frac{2}{\eta^{t}} \|\z_i(\mu)-\x_i^{t}\|^2 \\
&\leq& \min_{\mu\in[0,1]}
    (1-\mu)\,\hat{\phi}(\x_{1:n}^{t})
  + \mu\,\hat{\phi}(\x_{1:n}^*)
  + \sum_{i=1}^n \Big(\frac{2\mu^2}{\eta^{t}}-\frac{\mu(1-\mu)}{2}\Big)
    \|\x_i^{t}-\x_i^*\|^2 ~~.
\end{eqnarray*}
Finally, we replace the minimizer w.r.t. $\mu$ with a specific choice
$\mu=\frac{\eta^{t}}{4+\eta^{t}}$ and get,
\begin{eqnarray*}
\hat{\phi}(\x_{1:n}^{t+1}) &\leq&
 \frac{4}{4+\eta^{t}} \hat{\phi}(\x_{1:n}^{t})
+ \frac{\eta^{t}}{4+\eta^{t}} \hat{\phi}(\x_{1:n}^*) ~~.
\end{eqnarray*}
Therefore, the optimality gap of the surrogate $\hat{\phi}$ satisfies,
$$\hat{\phi}(\x_{1:n}^{t+1})-\hat{\phi}^*
\leq \frac{4}{4+\eta^{t}} \big(\hat{\phi}(\x_{1:n}^{t})-\hat{\phi}^*\big)
\leq \frac{4}{4+\epsilon} \big(\hat{\phi}(\x_{1:n}^{t})-\hat{\phi}^*\big)~~.$$
In summary, we get that the sequence $\hat{\phi}(\x_{1:n}^{t})$ converges
linearly to $\hat{\phi}^*$.

To conclude the section we would like to underscore the computational advantage of
using a surrogate loss with exact subgradient following over a dual ascent
algorithm {\em specifically} tailored for the problem. The dual proximal
gradient (DPG) is described and shortly analyzed in Appendix~\ref{dpg:app}. To
do so, we generated synthetic data of varying levels of variational sparsity as
follows. For each sparsity level, let $\mathcal{D}$ denote a distribution whose
density function satisfies $\mathcal{D}(\x)\sim \ee^{-\lambda \|\x\|}$ such
that its probability mass within the unit ball is equal to a prescribed fusion
probability.  We first sample a sequence of Boolean variables $\{\nu_i\}$ from
the Bernoulli distribution with a predefined probability which designated the
chance of fusing two consecutive vectors.  When $\nu_i=1$, we sampled dual
variables $\ba_i$ from $\mathcal{D}$ over the unit ball and set
$\x_{i+1}-\x_i=\0$. Otherwise, we sampled $\x_{i+1}-\x_i$ from $\mathcal{D}$
{\em outside} the unit ball and set $\ba_i\in\partial\|\x_{i+1}-\x_i\|$.  We
then constructed $\y_i$ based on the optimality condition,
$\y_i=\x_i-\ba_i+\ba_{i-1}$. This randomized generation process ensures that
$\x_{1:n},\ba_{0:n}$ are primal-dual optimal for $\y_{1:n}$.

We ran both DPG and the \moss\ for 100 iterations each. We then used the
resulting solution and merged two consecutive vectors when the $2$-norm of
their difference was smaller than $10^{-8}$. For each level of variational
sparsity we generated $100$ random sequences of vectors of dimension $d=100$
where each sequence is of length $n=100$. We measured the $2$-norm between the
solution found by the algorithms and the primal-dual optimal solution which was
used to generate the synthetic data. The results, shown in Figure~7, clearly
indicate that \moss\ is superior to DPG which despite the latter being tailored
for the problem. The gap between $\x^*_{1:n}$ and $\hat{\x}_{1:n}$ found by
\moss\ is insensitive to the fusion probability whereas DPG's performance
deteriorates as the level of sparsity increases. Equally, if not more
important, \moss\ exhibits much superior recovery rate of fusion events and in
fact seems to slightly improve with the level of sparsity. In contrast, DPG's
rate of recovery drastically falls to the point that at a sparsity level of
$70\%$ almost no fusion events were identified.
\begin{figure}[t]
\centering
\fbox{
\begin{subfigure}{.48\textwidth}
  \centering
  \includegraphics[width=\linewidth]{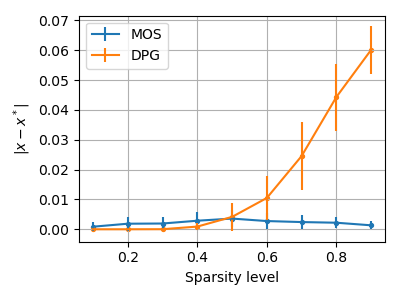}
\end{subfigure}%
\begin{subfigure}{.48\textwidth}
  \centering
  \includegraphics[width=\linewidth]{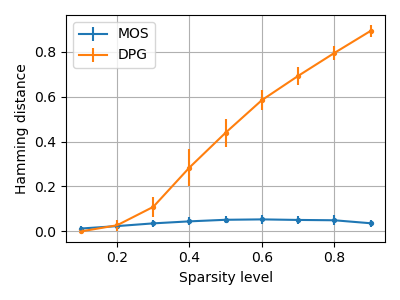}
\end{subfigure}
}
\caption{\moss\ vs. DPG for sparse multivariate discovery with
a $2$-norm penalty.}
\end{figure}

\subsection{Multivariate Regression With $\infty$-norm Variational Penalty}
The second vector fusion penalty we consider is the infinity norm of
the difference vectors $\x_i - \x_{i+1}$ which entails the problem of finding
the optimal solution of,
\beq{linfmv}
\frac12\sum_{i=1}^n \|\x_i-\y_i\|^2
+ \sum_{i=1}^{n-1} \| \x_{i+1} - \x_i\|_\infty ~~.
\eeq
We associate a slack variable with each difference vector $\xi_i$ and rewrite
the problem as,
\begin{eqnarray*}
\min_{\x_{1:n},\xib\geq 0} \;
\frac12\sum_{i=1}^n \|\x_i-\y_i\|^2 + \sum_{i=1}^{n-1} \xi_i &
\text{ s.t. } ~ & \|\x_{i+1} - \x_i\|_\infty \leq \xi_i ~~.
\end{eqnarray*}
We cast the problem as a gradient-based search for the optimum
$\xi_{1:n-1}^*$.

On each gradient step we solve a residual problem in $\x$ using the subgradient
following procedure with bounded variation from the previous section. Formally,
given any $\xi_{1:n-1}\geq 0$ we use the problem given by~\eqr{linfmv} to
define $\mathcal{L}(\xi_{1:n-1})$ as follows,
\begin{eqnarray*}
\mathcal{L}(\xi_{1:n-1}) = \min_{\x_{1:n}} \;
\frac12\sum_{i=1}^n \|\x_i-\y_i\|^2 + \sum_{i=1}^{n-1} \xi_i &
\text{ s.t. } ~ & \|\x_{i+1} - \x_i\|_\infty \leq \xi_i ~~.
\end{eqnarray*}
In order to compute the subgradient of $\mathcal{L}$, we introduce the dual
variables and write the constrained problem as,
\begin{equation*}
\mathcal{L}(\xi_{1:n-1}) = \min_{\x_{1:n}} \, \max_{\ba_{1:n-1}} \;
      \frac12\sum_{i=1}^n \|\x_i-\y_i\|^2
    + \sum_{i=1}^{n-1} \xi_i
    + \sum_{i=1}^{n-1} \ba_i^\top (\x_{i+1}-\x_i) - \xi_i \|\ba_i\|_1
~~.
\end{equation*}

From strong duality we can swap the order of min and max in the minimax problem to get,
\begin{eqnarray*}
\mathcal{L}(\xi_{1:n-1}) &=& \max_{\ba_{1:n-1}} \, \min_{\x_{1:n}} \;
  \frac12\sum_{i=1}^n \|\x_i-\y_i\|^2 + \sum_{i=1}^{n-1} \xi_i
  + \sum_{i=1}^{n-1} \ba_i^\top (\x_{i+1}-\x_i) - \xi_i \|\ba_i\|_1 \\
&=& \max_{\ba_{1:n-1}} \;
  \frac12\sum_{i=1}^n \|\y_i\|_2^2 - \|\y_i + \ba_i - \ba_{i-1}\|_2^2
  + \sum_{i=1}^{n-1} \xi_i \big(1 - \|\ba_i\|_1 \big) ~~.
\end{eqnarray*}
Denote by $\ba_{1:n-1}^*$ the maximizer of the above equation and let
$\vv{u}$ be the vector whose $i$'th coordinate is $u_i = 1 - \|\ba_i^*\|_1$.
For any $\xi_{1:n-1}'\geq 0$, we have
\begin{eqnarray*}
\lefteqn{\mathcal{L}(\xi_{1:n-1}) + \vv{u}^\top (\xi_{1:n-1}'-\xi_{1:n-1}) =} \\
& & \displaystyle
\frac12\sum_{i=1}^n \|\y_i\|_2^2 - \|\y_i + \ba_i^* - \ba_{i-1}^*\|_2^2
+ \sum_{i=1}^{n-1} \xi_i' \big(1 - \|\ba_i^*\|_1 \big) \leq \mathcal{L}(\xi_{1:n-1}')
~ .
\end{eqnarray*}

Therefore $\vv{u}\in \partial\mathcal{L}(\xi_{1:n})$, namely, it is a
subgradient of $\mathcal{L}$ at $\xi_{1:n-1}$. Using strong duality again, we
can obtain the optimal dual variables by first solving the primal problem and
then infer the dual variables $\ba_{:}$ from optimality conditions. Given
$\xi_{1:n-1}$ the primal is separable and we thus can apply the subgradient
following from \ref{barrier} to obtain both the optimal primal and dual
variables. The slack variables $\xi_{1:n-1}$ are updated using gradient
stepping. To do so, we derive  upper bounds such that $\xi_i^* \leq D_i$ and
then apply AdaGrad~\cite{duchi2011adaptive} with an initial per-coordinate
learning rate of $D_i/\sqrt{2}$. The derivation of the upper bounds is given in
Appendix~\ref{upperb:app}.

To conclude the section, we describe the results of experiments with datasets
generated analogously to the ones constructed in previous section, replacing
$\|\cdot\|$ with the $1$-norm. As before we compare the results with a
specialized DPG procedure tailored for the problem. Each iteration of DPG now
takes $\mathcal{O}(nd\log{d})$ as we need to perform a projection onto the unit
ball w.r.t the $1$-norm. In contrast, the \moss\ would take
$\mathcal{O}(dn\log{n})$. With $d=100$ and $n=100$, the per-iteration cost is
morally the same. The results are summarized in Figure~8. The advantage of
\moss\ over DPG both in terms of recovery of fusion events (variational
sparsity) without any false discovery and in terms of proximity to the optimal
solution is strikingly apparent.
\begin{figure}[t]
\centering
\fbox{
\begin{subfigure}{.45\textwidth}
  \centering
  \includegraphics[width=\linewidth]{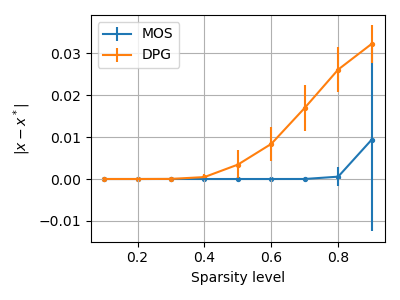}
\end{subfigure}%
\begin{subfigure}{.45\textwidth}
  \centering
  \includegraphics[width=\linewidth]{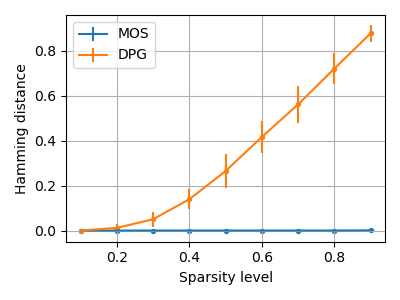}
\end{subfigure}
}
\caption{\moss\ vs. DPG for sparse multivariate discovery using the
the $\infty$-norm .}
\end{figure}

\section{Empirical Study} \label{empirical}
Prior to recapping the main results and concluding, we present in this section
an empirical study that underscores the potential of the high-order \moss. The
experiments do not pit one version of algorithm versus another and claim
superiority of a particular version, on the contrary. We describe experiments
with three different financial tickers of different temporal characteristics in
order to exhibit the merits of different versions in the light of different
stochastic settings. The three datasets that we experimented with are SPX, VIX,
and SFXRSA which we describe shortly below.

\begin{figure}[!t]
\label{findata}
\centering
\begin{subfigure}{.52\textwidth}
  \centering
  \includegraphics[width=\linewidth]{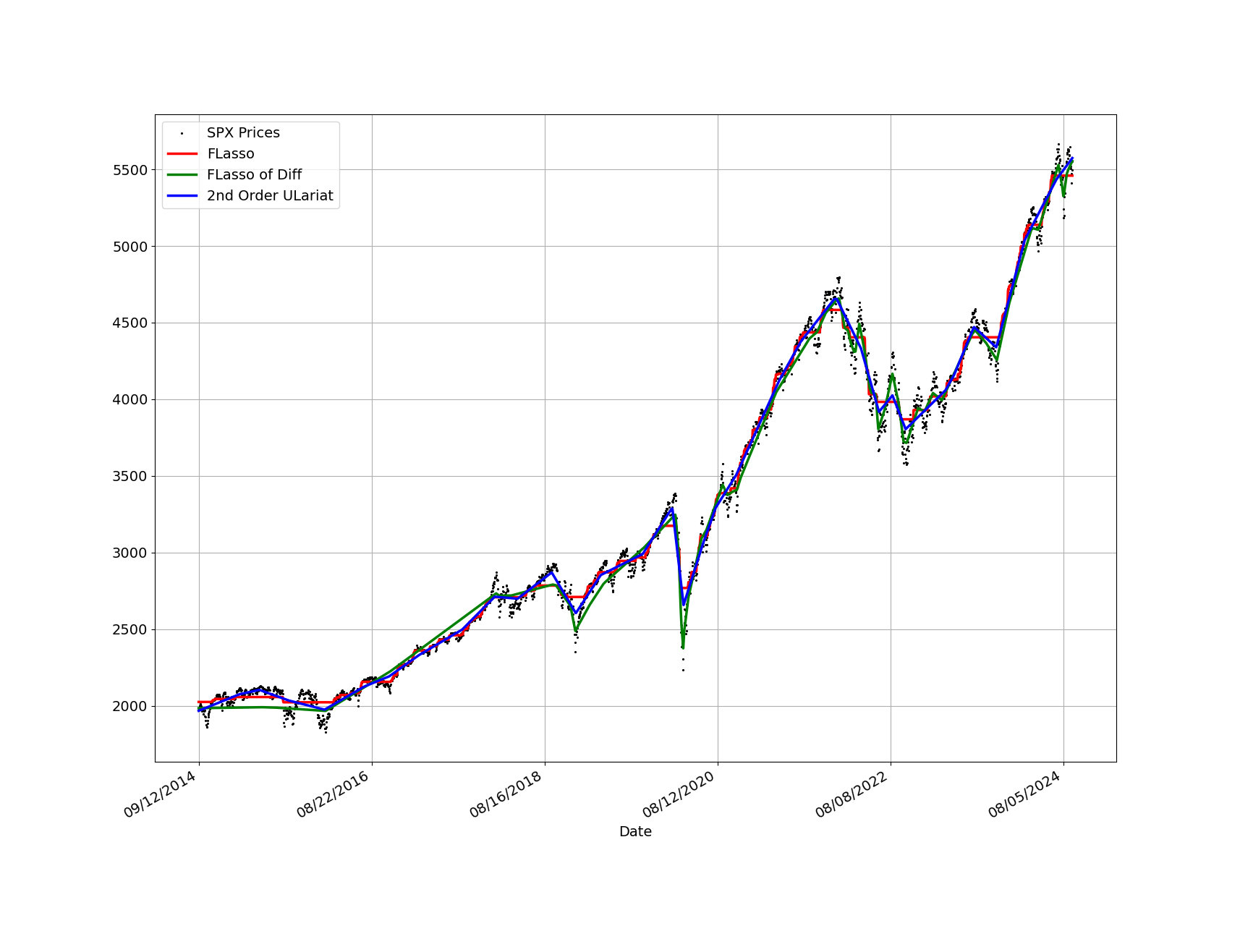}
\end{subfigure}%
\hspace{-0.96cm}
\begin{subfigure}{.52\textwidth}
  \centering
  \includegraphics[width=\linewidth]{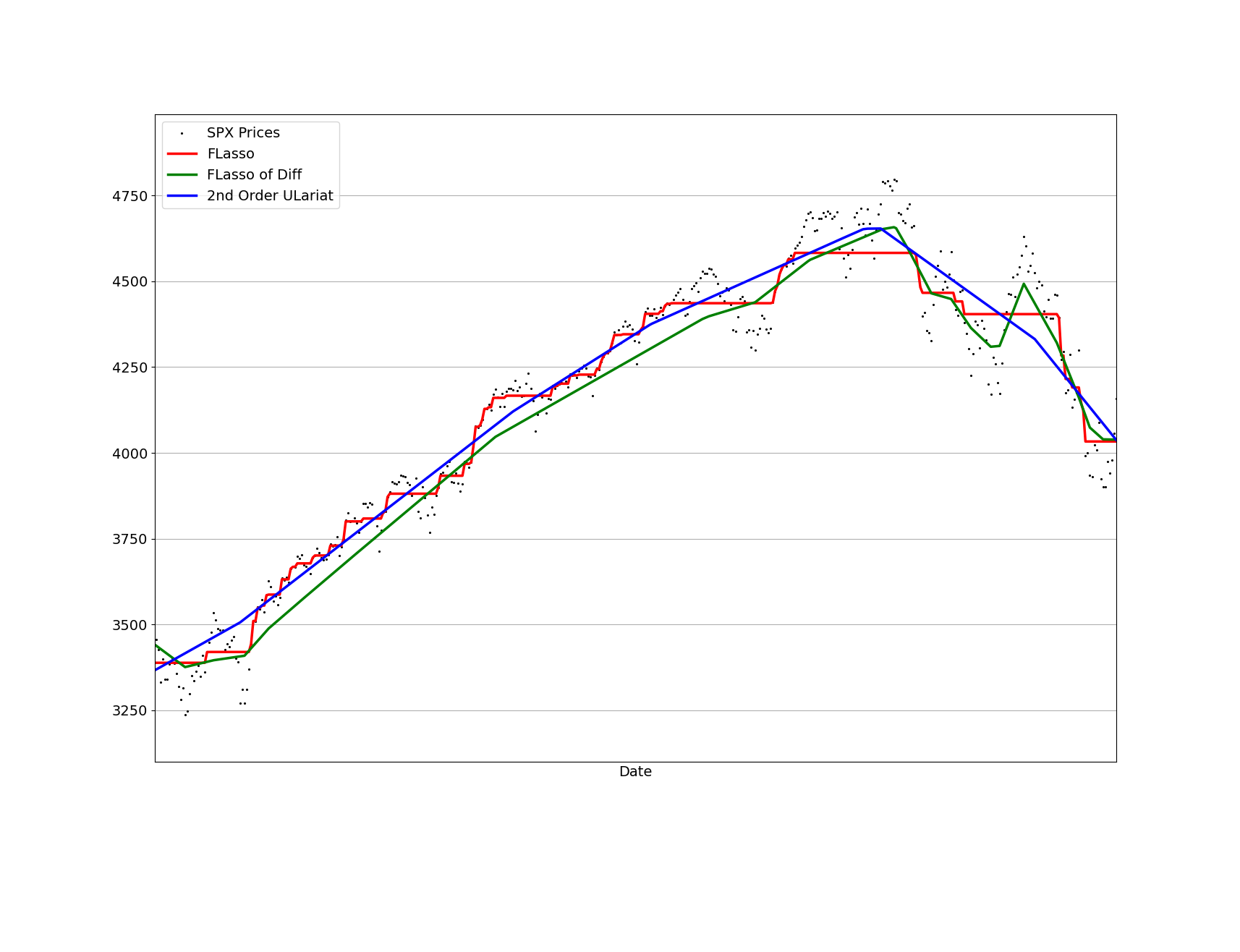}
\end{subfigure}
\centering
\begin{subfigure}{.52\textwidth}
  \centering
  \includegraphics[width=\linewidth]{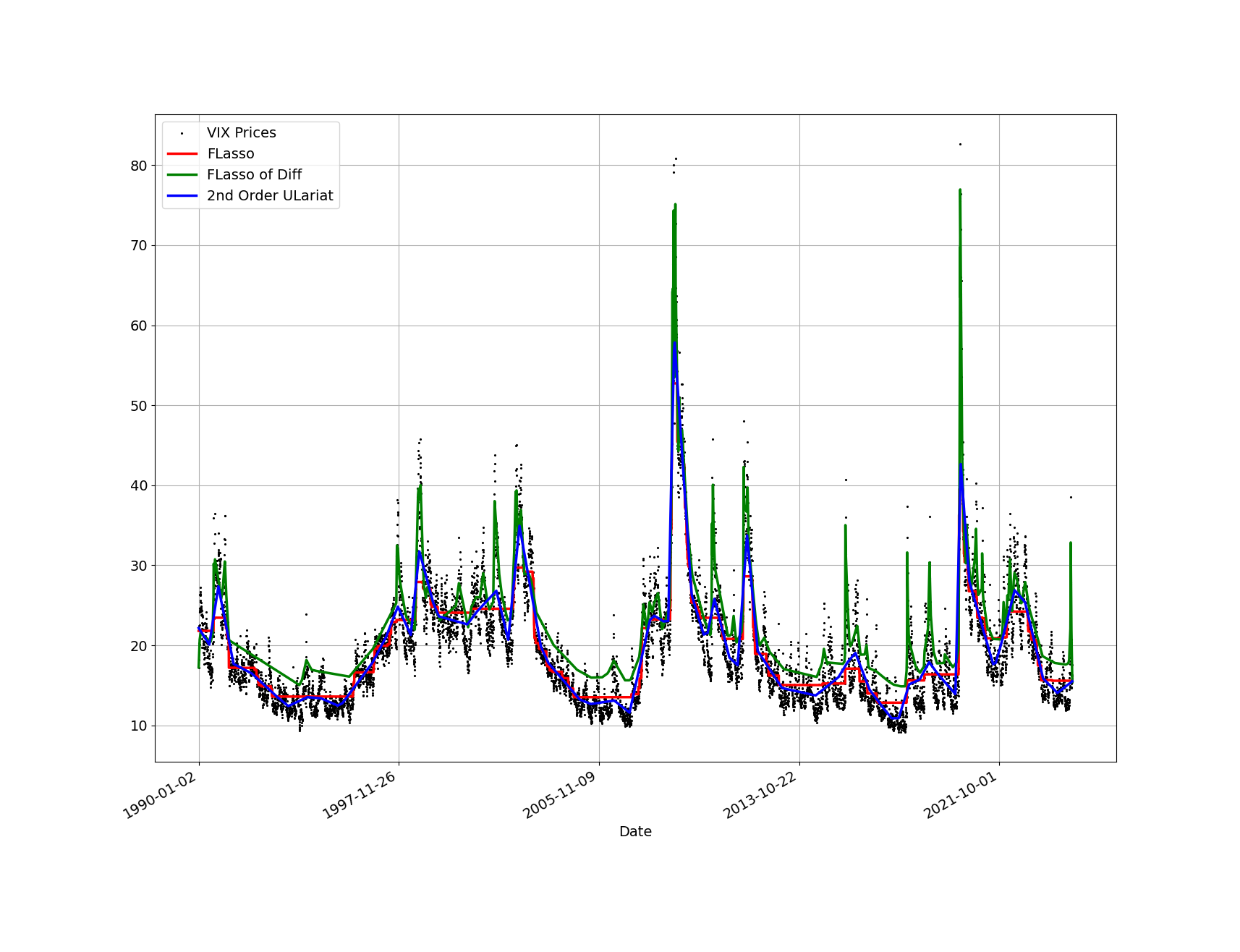}
\end{subfigure}%
\hspace{-0.96cm}
\begin{subfigure}{.52\textwidth}
  \centering
  \includegraphics[width=\linewidth]{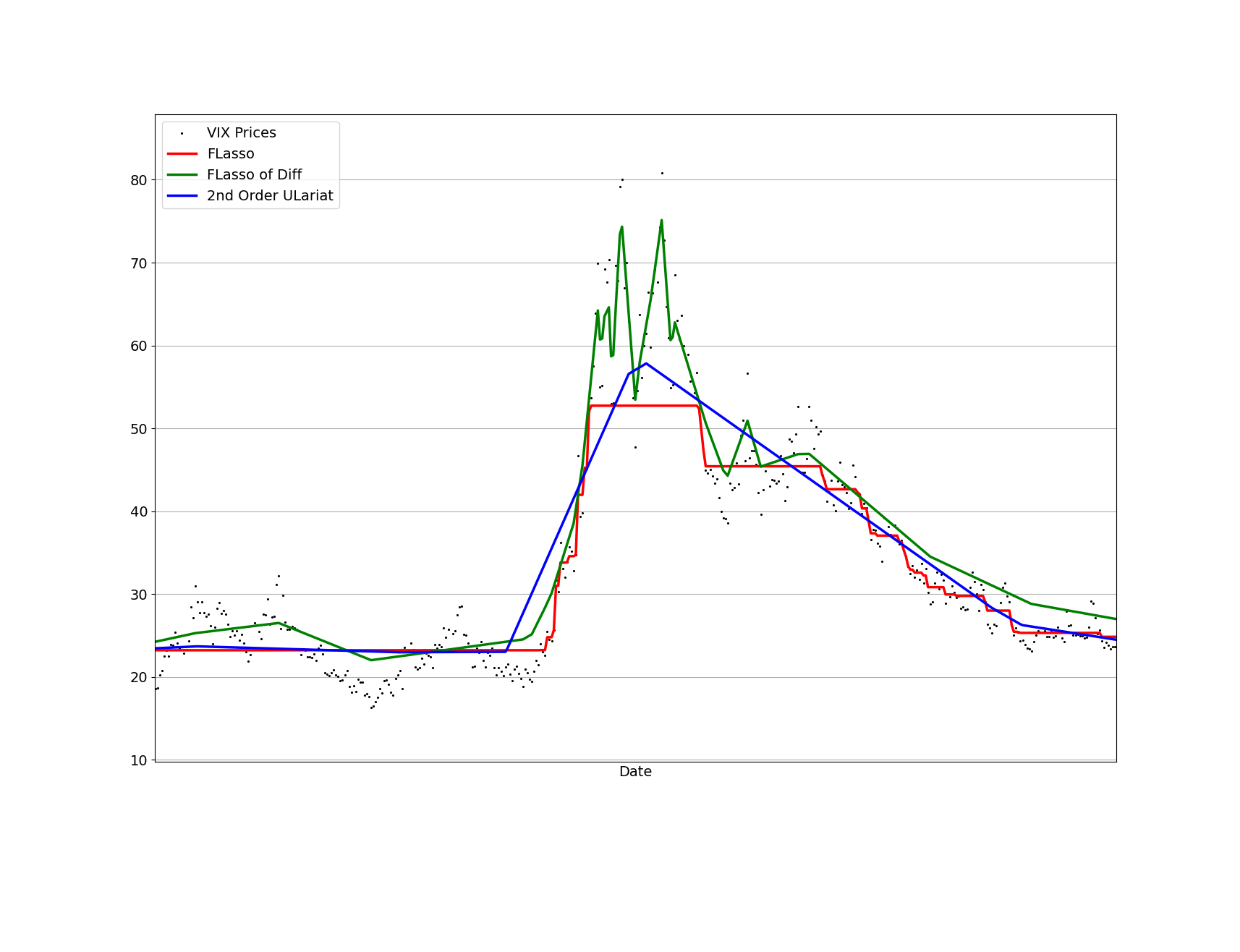}
\end{subfigure}
\centering
\begin{subfigure}{.52\textwidth}
  \centering
  \includegraphics[width=\linewidth]{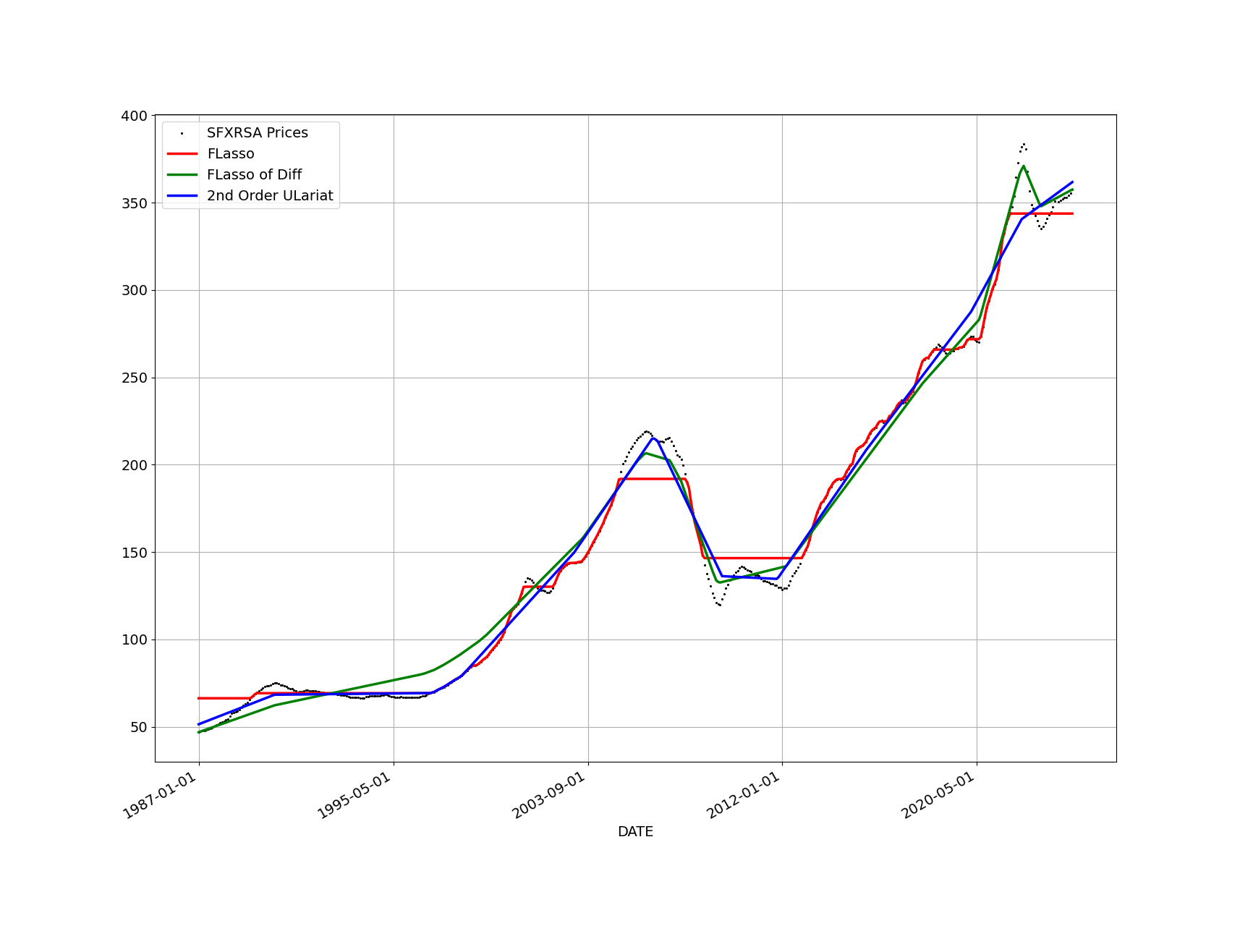}
\end{subfigure}%
\hspace{-0.96cm}
\begin{subfigure}{.52\textwidth}
  \centering
  \includegraphics[width=\linewidth]{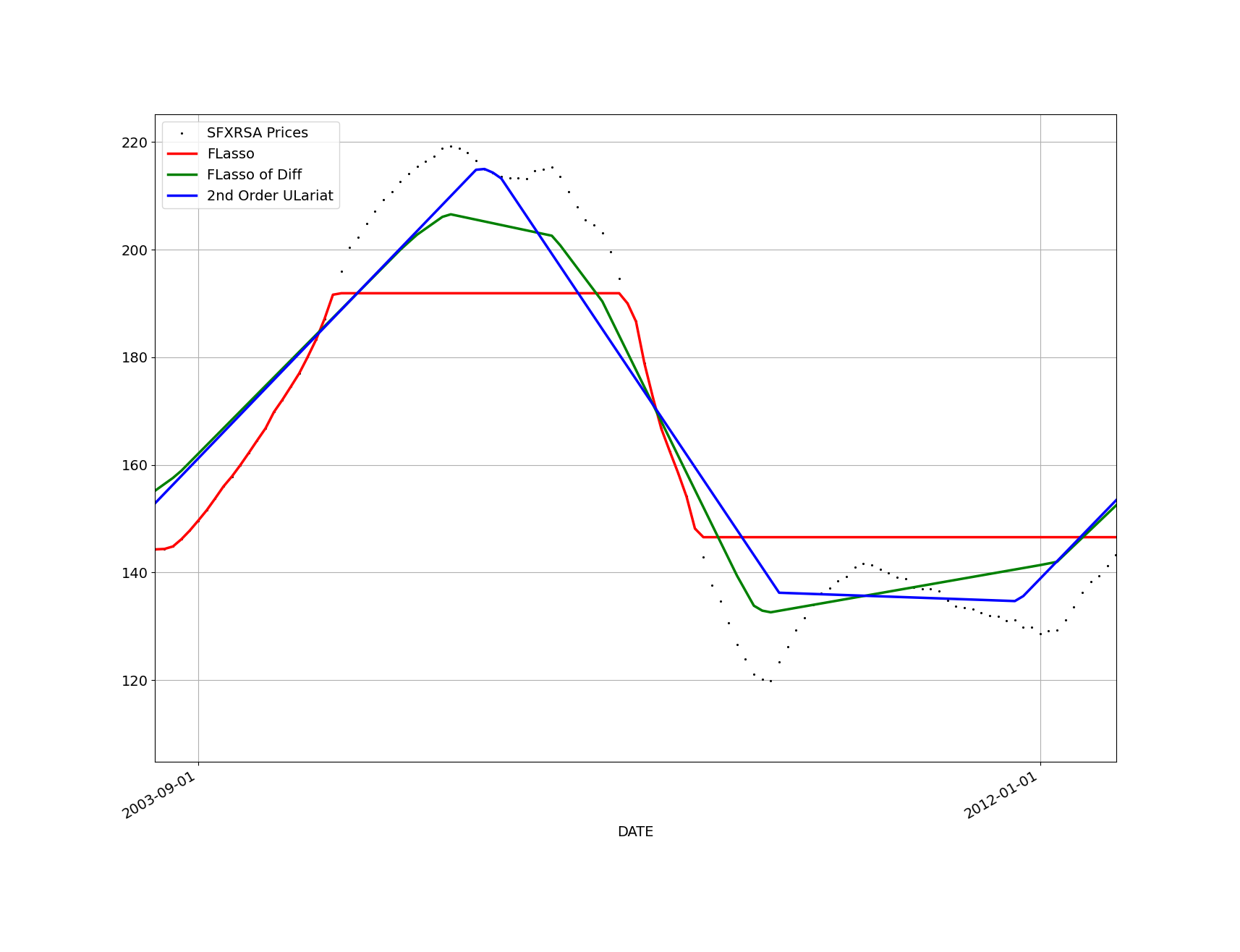}
\end{subfigure}
\label{findata}
\vspace{-0.1cm}
\caption{Comparison of Fused Lasso and \moss\ on financial indices.}
\label{findata}
\end{figure}

SPX is the ticker symbol for the Standard and Poor's 500, a stock market
index tracking the stock performance of 500 of the largest companies listed on
stock exchanges in the United States. It is one of the most commonly followed
equity indices and includes approximately 80\% of the total market
capitalization of US public companies. SPX is the case study of many research
papers and has garnered numerous stock market models.

VIX is the ticker symbol for the Chicago Board Options Exchange's CBOE
Volatility Index, a popular measure of the stock market's expectation of
volatility based on S\&P 500 index options. It is calculated and disseminated on
a real-time basis by the CBOE, and is often referred to as the fear index or
fear gauge. The VIX index tends to oscillate between periods of low volatility
with a small magnitude of it and changes with spike-like transient periods. The VIX
is rather difficult to model.

SFXRSA is a S\&P Case-Shiller Home Price Index that measures the price level of
existing single-family homes in San Francisco. The index reflects the average
change in home prices  in terms of percentage changes in housing market prices
given a constant level of quality. Changes in the types and sizes of houses or
changes in the physical characteristics of houses are specifically excluded
from the calculations to avoid incorrectly affecting the index value. It is
thus a slowly varying signal with some seasonal effect.

The three particular variants that we examined were the original Fused Lasso,
and \moss\ with a second order penalty equivalent for penalizing the total
variation of the discrete second derivative. For comparison we also ran the
Fused Lasso on the sequence of differences $y_{i+1} - y_i$. The latter mimics
the $2$'nd order \moss. The discrete derivative suppresses the low frequency
components of the original sequence, obviating completely the zero frequency
component.  We tuned $\lambda$ for the three variants below so that their
residual error of the entire series of each financial index is the {\em same}.
Thus, the different behavior of each filtering approach reflects their
characteristics and merits.  As written above, the goal of the comparison is by
no means to show superiority or a universal advantage of the nascent \moss\
over the Fused Lasso or other methods mentioned in this paper. Indeed, there
are settings in which the Fused Lasso would constitute the most viable data
analysis tool, for instance during prolonged periods of economic stagnation.

The results for the three financial indices are depicted on the left column of
Fig.~\ref{findata}. The three procedures seem to capture the overall tendency
of the three different indices. The approximated indices generated by Fused
Lasso are, as one would expect, piecewise constant. Overall, the approximation
does not seem to provide an aesthetic refinement of the original data. We would
henceforth more closely examine solely the performance of the Fused Lasso on the
sequence of difference and the \moss\ on the original data. Upon a first
examination it is apparent that the Fused Lasso on the difference sequences is
capable of capturing transient periods during which there is typically a sharp
upturn followed by a similar or less pronounced downturn. However, there also
seems to be periods of systematic bias within the approximated sequences. Since
the finite derivatives of the original indices eliminates zero
frequency effects, the biased estimates do not affect the overall error.

To further qualitatively examine the results, we provide on the right column of
Figure~\ref{findata} zoomed-in snippets of an upturn-downturn period. These
snippets are provided for visualization purposes and were {\em not} handpicked.
All three variants were provided the entire sequence, not the particular
snippets we visualize. Traversing the figures from top hand-side to bottom
hand-side, for SPX we see that the Fused Lasso's approximation is
over-fragmented and does not provide a faithful approximation of the
up-then-down characteristics. The Fused Lasso on the sequence of price changes
does capture nicely the temporal wedge shape but it oversubscribes to small
changes on occasion. The $2$'nd order variant of \moss\ nicely captures the
wedge characteristics with about $4$ piecewise linear segments.

By construction and definition of the VIX index, it is more volatile than SPX.
In the middle row we examine a snippet of the VIX index during a morally
up-down period albeit with further nuanced changes that could be significant
for a financial data analyst. The approximation of the Fused Lasso on the price
changes in comparison with $2$'nd order \moss\ underscores a natural trade-off.
The latter approximates VIX by morally 4 linear segments, a period without a
substantial change that is approximated as a sequence of fixed values, followed
by a single segment of upturn, concluding with 3 segments of linear decrease in
the index. It thus faithfully captures the overall trend. In contrast, the
Fused Lasso on price changes still captures the nature of the sub-sequence is
more fragmented and sensitive to local changes albeit resulting with the same
approximation error as the \moss\ due to the respective choices of $\lambda$.
It thus provides a more pinpointed tool during high volatility periods that are
absent from the overall market trend. Lastly, for the real-estate index SFXRSA,
all three variants seem to capture quite faithfully the overall behavior of the
slowly changing real-estate market in San Francisco.

We would like to note that we can possibly gain the best of all approaches with
a further generalized version of \moss\ with composite variational penalty
consisting of two terms such as,
$$
  \lambda_1 |x_{i+1}-x_{i}| + \lambda_2 |x_{i-1} - 2 x_i + x_{i+1}| ~~.
$$
We leave the derivation and analysis of \moss\ with composite variational
penalties to future research.

\section{Concluding Remarks}
This work (subjectively) makes the following novel contributions:
{\small
\begin{itemize}
  \item Provides a unified approach for sequence approximation w.r.t.
  Bregman divergences with general total variation penalties.

  \item Describes and analyzes a novel subgradient following procedure,
  \moss, with succinct code that can be readily used with existing and new
  variation penalties.

  \item \moss\ entertains the same time complexity as known algorithms, each of
  which designed (by different authors) for a concrete variational penalty.

  \item Introduces and analyzes a multivariate generalization where each
  element of the sequence is a vector. The base case of squared $2$-norm as the
  variational penalty of two consecutive vectors bears a closed form for
  subgradient following.

  \item Derives iterative algorithms for sparse multivariate total variation
  using the $2$-norm or the $\infty$-norm by employing the norm-squared
  version as surrogate.

  \item Describes and analyzes nascent high-order \moss\ with poly-time
  subgradient following.
\end{itemize}
}

There are several possible extensions of the framework presented in this
paper. The univariate penalties are readily extendable to non-symmetric
counterparts and the addition of insensitivity regions. For instance,
a generalization the symmetric penalty of the fused lasso amounts to,
$$
  \lambda_i^\mathsf{R}\big[x_{i+1}-x_i-\epsilon_i^{\mathsf{R}}\big]_+ \,+\,
  \lambda_i^\mathsf{L}\big[x_{i}-x_{i+1}-\epsilon_i^{\mathsf{L}}\big]_+
  ~~.
$$
By choosing $\lambda_i^\mathsf{R}=\lambda_i^\mathsf{L}=\lambda_i$ and
$\epsilon_i^{\mathsf{R}}=\epsilon_i^{\mathsf{L}}=0$ we obtain the fused
lasso's penalty. To encompass this version we simply need to compute
$\partial g_i^*$ from which we derive a new mapping
$\ix \mapsto \ia_i(\ix)$. The subgradient procedure uses a similar form
for the mapping given in \eqr{aix} though the introduction of insensitivity
regions to the left and/or the right of the loci where $x_i=x_{i+1}$ would
introduce two additional inflection points. The resulting procedure for
constructing the subgradient would require at most $\oo(n\log n)$ time.

We limited our discussion to sequential penalties which can described as a
chain of local connectivities $x_1 \leftrightarrow x_2 \leftrightarrow
\ldots x_i \leftrightarrow x_{i+1} \ldots $ forming a left-to-right
dependency graph.  Kolmogorov et. al~\cite{kolmogorov2016total} examined a
setting where the total variation penalties form a tree.  Denote the
variation dependency graph by $G = (V, E)$ where $V = [n]$ and $(i,j)\in E$
if there exists a penalty term for the variables $x_i$ and $x_j$. Using the
graph representation, the penalized problem takes the form of,
\begin{equation*}
U = \min_{\x}\: \sum_{i} \frac12 (x_i - y_i)^2 +
  \sum_{(i,j)\in E} |x_i - x_j| ~~.
\end{equation*}
In settings for which strong duality holds the above problem can be
re-parameterized as,
\begin{eqnarray*}
U &=& \max_{\|\ba\|_\infty\leq 1}\, \min_{\x}\:
  \frac12 \sum_{i=1}^n (x_i - y_i)^2 
  + \sum_{(i,j)\in E} \alpha_{i,j} (x_i - x_j) ~~.
\end{eqnarray*}
When $E$ defines a tree the subgradient following procedure can utilize
the form described in~\cite{kolmogorov2016total} albeit increasing the
time complexity to $\oo(n^2)$. 

\paragraph{Acknowledgments}
We would like to thank Tomer Koren and Yishay Mansour for fruitful
discussions. The short discussion above is motivated by conversations with
the late Prof. Tali Tishby who is missed dearly. This manuscript was written
without the usage of AIML tools. 

\bibliographystyle{plain}
\bibliography{bib}

\appendix

\section{Improved Barrier Lariat using Skiplists} \label{skipl:app}
We store each $\ia_i(\cdot)$ as a linked-list data structure of implicit change
points (vertices) and explicit linear intervals connecting two consecutive
vertices. A vertex $(v_j,\ia_i(v_j))=(v_j,\alpha_j)$ represents a point where
$\ia_i(\cdot)$ changes slope. An edge $(d_j,s_j)$ stores the length and slope
of the line segment between $v_j$ and $v_{j+1}$, thus we have,
$
  d_j = v_{j+1} - v_j \,;\, s_j = (\alpha_{j+1} - \alpha_j) / d_j ~~.
$
This representation allows lazy evaluations in a similar fashion to the
fused lasso. The \kvn\ step does not change the lengths of intervals
except for the new segment encapsulating the zero crossing. Since the length
of all other line segments remain intact this edge-based representation
facilitates an efficient implementation.

By construction of $\ia_{i+1}$ we know that $\exists j$ for which
$v_j=z_{i+1}-\lambda_{i+1}$ and $\ia_{i+1}(z_{i+1}-\lambda_{i+1})=0$. This
implies that in an explicit representation of $\ia'_i(\cdot)$ we would need to
update vertex $j$,
$$
  (z_{i+1}-\lambda_{i+1},0)
    ~~\mapsto~~
  (z_{i+1}-\lambda_{i+1},z_{i+1}-\lambda_{i+1}-y_{i+1}) ~~.
$$
In the explicit view of the lazy representation of the ensuing edge, the slope
$s_j$ of the segment $[v_j,v_{j+1}]$ becomes $s_j+n-i$. We next find the zero
crossing point of $\ia'_i(\cdot)$ by traversing the list of edges towards zero
and locating the index $k$ for which $\alpha_k\leq 0$ and $\alpha_{k+1} > 0$.
In case such an index does not exist the zero of $\ia'_i(\cdot)$ is right of
its rightmost vertex. Once we identify the enclosing interval, we calculate
$z_i$ by solving a linear equation from $(v_k,\alpha_k)$ and
$(v_{k+1},\alpha_{k+1})$. We next \kvn\ the list by inserting two vertices,
$(z_i-\lambda_i,0)$ and $(z_i+\lambda_i,0)$, between $k$ and $k+1$,
and update $d_k$ accordingly. This sequential traversal of the list
in the worst-case would require linear time and thus total run time would
remain intact at $\oo(n^2)$.

In order to employ a more efficient procedure for finding zero crossings, we
make use of a skip-list data structure~\cite{pugh1990skip}. A skip-list is an
extension of a linked list with multiple layers and with which the expected
search time is $\oo(\log{n})$. The skip-list has ideally in our setting
$\oo(\log{n})$ layers where layer $k$ contains all vertices whose indices
modulo $2^k$ are zero. The search procedure starts from the top layer and
proceeds downwards once we reach the interval encapsulating zero. In practice,
a skip-list is a randomized data structure where each vertex appears in a
higher layer with probability of $\frac12$ to support efficient insertions.

We follow the same lazy representation as above. We store
$(d_j^k,s_j^k)=(v_{j+1}^k-v_j^k,s_j^k)$ for edge $j$ at layer $k$. We also add
a precursor vertex at the beginning of each layer. This vertex contains the
smallest possible value of $\{v_j\}$ and serves as a boundary point. To
construct $\ia'_i(\cdot)$ we update $\alpha_0\gets\alpha_0+v_0-y_{i+1}$. Then,
we follow the search procedure above until reaching the bottom layer. We
calculate $z_i$ at the bottom layer by solving the same linear equation as
above. During the insertion of vertex $(z_i-\lambda_i,0)$, we flip a coin to
decide whether it should appear in a higher layer. Note that along the
searching path, we have calculated $(v_j^k,\alpha_j^k)$ and
$(v_{j+1}^k,\alpha_{j+1}^k)$ per layer which contains the locus of the zero
crossing. Thus, whenever a vertex should appear in a prior (higher) layer, we
use the corresponding values for updating the connecting edges. We perform an
equivalent update for $(z_i+\lambda_i,0)$.  Last, we update the $v_0\gets
v_0-\lambda_i$. The end result is an $\oo(n\log{n})$ {\em amortized} time
algorithm.

\section{Dual Proximal Gradient} \label{dpg:app}
We give here further details on the dual proximal gradient method (DPG) used in
the empirical studies. DPG merits further investigation on its own and to the
best of our knowledge the short analysis with general norm penalties below was
not derived in previous research.

Let $\|\cdot\|$ denote a $p$-norm and $\|\cdot\|_*$ its dual norm. We
are interested in solving the following problem,
$$
  \min_{\x_{1:n}} ~
    \frac12 \sum_{i=1}^n \big\|\x_i-\y_i\big\|_2^2 \;+\;
    \sum_{i=1}^{n-1} \big\| \x_{i+1} - \x_i\big\| ~~.
$$
The dual of this problem is,
\begin{equation*}
  \max_{\ba_{0:n}} ~
  \sum_{i=1}^n \big\|\y_i\big\|_2^2 - \big\|\y_i + \ba_i - \ba_{i-1}\big\|_2^2
  ~\st~ \big\|\ba_i\big\|_* \leq 1 ~~.
\end{equation*}
For boundary conditions we set $\ba_0 \equiv \ba_n \equiv \0$.
Denote by $B_* = \{\x : \|\x\|_* \leq 1\}$ the unit ball w.r.t the dual norm.
Let $\Pi_{B_*}(\cdot)$ denote the projection operator onto the unit ball.
In each iteration of dual proximal gradient,
$$
\ba_i \,\gets\, \Pi_{B_*}
  \Big(\ba_i - \eta (\ba_i-\ba_{i-1}+\y_i + \ba_i-\ba_{i+1}-\y_{i+1}) \Big)
~~.
$$
To underscore the advantage of the sub-gradient following method,
let us briefly examine the condition number of the dual problem.
Denote by $\m{D}$ the difference matrix
$$\m{D} = \begin{bmatrix}
    -1 & 1 & 0 & \dots  & 0 & 0 \\
    0 & -1 & 1 & \dots  & 0 & 0 \\
    \vdots & \vdots & \vdots & \ddots & \vdots \\
    0 & 0 & 0 & \dots  & -1 & 1
\end{bmatrix} 
\in \R^{n-1\,\times\,n}  ~~.$$
The condition number of the dual problem is the same as that of
$\m{D}\m{D}^\top$. If we choose $\ba$ such that
$\alpha_i = \frac{{(-1)}^i}{\sqrt{n-1}}~,$
we have
$\|\m{D}^\top\ba\|_2^2 = \ba^\top\m{D}\m{D}^\top\ba = 4 - \frac{2}{n-1}$.
Assume without loss of generality that $n$ is even and let us choose next 
$$\alpha_i = \frac{\frac{n}{2} - |i - \frac{n}{2}|}{\sqrt{n^3/12}}~~,$$
for which we have $\|\ba\|_2\geq 1$ and $\ba^\top\m{D}\m{D}^\top\ba = 12/n^2$.
Therefore, for $n\geq 3$ the largest eigenvalue of $\m{D}\m{D}^\top$ is
at least $3$ whereas the smallest eigenvalue is at most $\mathcal{O}(n^{-2})$.
Therefore, the condition number of the dual problem is $\Omega(n^2)$ which
renders dual ascent algorithms, and in particular DPG, slow to converge.

\section{Upper Bounds on Slack Variables} \label{upperb:app}

Denote the optimum of the original problem by $\x_{1:n}^*$,
we claim that $\|\x_{i+1}^*-\x_i^*\|_\infty\leq \|\y_{i+1}-\y_i\|_\infty$
and thus we can set $D_i=\|\y_{i+1}-\y_i\|_\infty$.
If $\|\x_{i+1}^*-\x_i^*\|_\infty = 0$, the bound clearly holds.
Otherwise let us define,
$$S = \{j : |x_{i+1,j}^*-x_{i,j}^*|=\|\x_{i+1}^*-\x_i^*\|_\infty\}~~.$$
Consider $j\in S$ and let $s_{i,j} = \sign(x_{i+1,j}^*-x_{i,j}^*)$.
From the optimality condition, there exists $\ba_{1:n-1}$ such that
\begin{eqnarray*}
x_{i+1,j}^*-x_{i,j}^* &=& (y_{i+1,j}+\alpha_{i+1,j}-\alpha_{i,j}) -
  (y_{i,j}+\alpha_{i,j}-\alpha_{i-1,j}) \\
&=& y_{i+1,j}-y_{i,j} + \alpha_{i+1,j}-2\alpha_{i,j}+\alpha_{i-1,j} ~~.
\end{eqnarray*}
In addition, we have $s_{i,j}\alpha_{i,j}\geq 0$.
Multiplying both side by $s_{i,j}$ we get
\begin{eqnarray*}
|x_{i+1,j}^*-x_{i,j}^*| &=& s_{i,j}(y_{i+1,j}-y_{i,j})
+ s_{i,j}\alpha_{i+1,j}+s_{i,j}\alpha_{i-1,j} - 2|\alpha_{i,j}| \\
&\leq& |y_{i+1,j}-y_{i,j}|+|\alpha_{i+1,j}|+|\alpha_{i-1,j}|-2|\alpha_{i,j}| ~~.
\end{eqnarray*}
Summing over $j\in S$ and dividing by $|S|$ we get,
\begin{eqnarray*}
  \big\|\x_{i+1}^*-\x_i^*\big\|_\infty
    & \leq &
      \frac{1}{|S|}\Big(\sum_{j\in S}|y_{i+1,j}-y_{i,j}| +
      \|\ba_{i+1}\|_1 + \|\ba_{i-1}\|_1 - 2\Big) ~
   \leq ~ \|\y_{i+1}-\y_i\|_\infty ~~,
\end{eqnarray*}
where for obtaining the second inequality we used the fact that for
$j\notin S: \alpha_{i,j}=0$ and $\|\ba_i\|_1 = 1$ since
$\|\x_{i+1}^*-\x_i^*\|_\infty \neq 0$.

\section{Iterative Algorithm for High-Order Variational Penalties} \label{high_order:app}
Note that the following modified smooth problem can be solved in linear time,
\begin{equation*}
\frac12\|\x-\y\|^2
+ \sum_{i=1}^{n-2} \lambda_i\big| (x_{i+2} - x_{i+1}) - (x_{i+1} - x_i)\big|^2 ~~.
\end{equation*}
Since the problem is differentiable everywhere, the optimality condition
amounts to a linear system $\m{D}\x = \y$ where $\m{D}$ is a five-diagonal matrix.
The system can be solved using Gauss elimination to reduce the matrix
to an upper triangular one in linear time. Once triangulated, $\x$ is simply
read out from the resulting system.  Then, similar to Algorithm~1, we iteratively
replace the non-smooth objective by its quadratic upper bound.
Namely, on iteration $t$, we obtain an exact solution from,
$$\x^{t+1} = \displaystyle
  \argmin_\x
    \|\x-\y\|^2 +
    \sum_{i=1}^{n-2}
    {\big|(x_{i+2}-x_{i+1}) - (x_{i+1}-x_i)\big|^2} \,/\, {\rho_i^{t}}
~~,$$
where
$\rho_i^{t} = \displaystyle
\max\big(\big|(x_{i+2}^{t} - x_{i+1}^{t}) - (x_{i+1}^{t} - x_i^{t})\big|,\, \epsilon\big)$.

The line of proof as above can be carried out except that now we have,
$$
  |(x_{i+2} -2 x_{i+1} +x_i) - (x'_{i+2} -2 x'_{i+1} +x'_i)|^2
   \leq 4 \left(|x_{i+2}-x'_{i+2}|^2 + 2|x_{i+1}-x'_{i+1}|^2 + |x_i-x'_i|^2\right)
~~.$$
Summing over $i$ we end up with a factor of 16 instead of 4.
This procedure can be generalized to $k$'th order finite difference
penalties. Concretely, let us define the matrix by $\tilde{\m{D}}$
the discrete differentiation matrix with zero padding,
$$\tilde{\m{D}} = \begin{bmatrix}
    -1 & 1 & 0 & \dots  & 0 & 0 \\
    0 & -1 & 1 & \dots  & 0 & 0 \\
    \vdots & \vdots & \vdots & \ddots & \vdots \\
    0 & 0 & 0 & \dots  & -1 & 1 \\
    0 & 0 & 0 & \dots  &  0 & 0\\
\end{bmatrix} 
\in \R^{n\,\times\,n}  ~~.$$
We also employ this matrix sans the last row in Appendix~\ref{dpg:app}.
The penalized $k$'th order finite differences we care to solve is,
$$
\argmin_\x \frac12\big\|\x-\y\big\|^2 +
  \big\| \big[\m{\tilde{D}}^k \x\big]_{1:n-k} \big\|_1 ~~. 
$$
As for runtime, we ``pay'' a factor of $4^{k}$ for the number of
iterations required for convergence where each iteration takes $\mathcal{O}(k^2 n)$
time for solving the linear system. In comparison, when using the dual proximal
gradient (DPG) method, the iteration complexity grows as
$\mathcal{O}(\gamma^k)$ where $\gamma=\Omega(n^2)$ is the condition number of
the first order problem. This yields a substantially inferior time complexity
of $\oo(n^{2k})$ for DPG. Pseudocode for $k>0$ is provided below.
\begin{figure}[ht]
\begin{center}
\begin{minipage}{.60\linewidth}
\begin{algorithm}[H]
\label{alg:high-ordrer}
\caption{$k^{\text{th}}$-Order \moss}
\begin{algorithmic}
\STATE{\bf initialize} $\x^{0} = \y$
\FOR{$t = 1\ \textbf{to}\ T$}
\smallskip
\STATE $\m{H}^t = \begin{bmatrix}
		\diag\bigg(\bigg[\Big|
\big[\m{\tilde{D}}^k\,\x^t\big]_{1:n-k} \Big|\bigg]_\epsilon\bigg)^{-1/2} & \m{O}_{n-k\times k} \\
		\m{O}_{k\times n-k} & \m{O}_{k\times k} \\
\end{bmatrix}$
\smallskip
\STATE $\x^{t+1} = \displaystyle \argmin_\x \frac12\|\x-\y\|^2
+ \frac12 \| \m{H}^t\,\m{\tilde{D}}^k\,\x \|^2
$
\ENDFOR
\STATE \textbf{return} $\vv{x}$
\end{algorithmic}
\end{algorithm}
\end{minipage}
\end{center}
\end{figure}

\begin{figure*}[t]
\centerline{{\fbox{\includegraphics[width=0.825\textwidth]{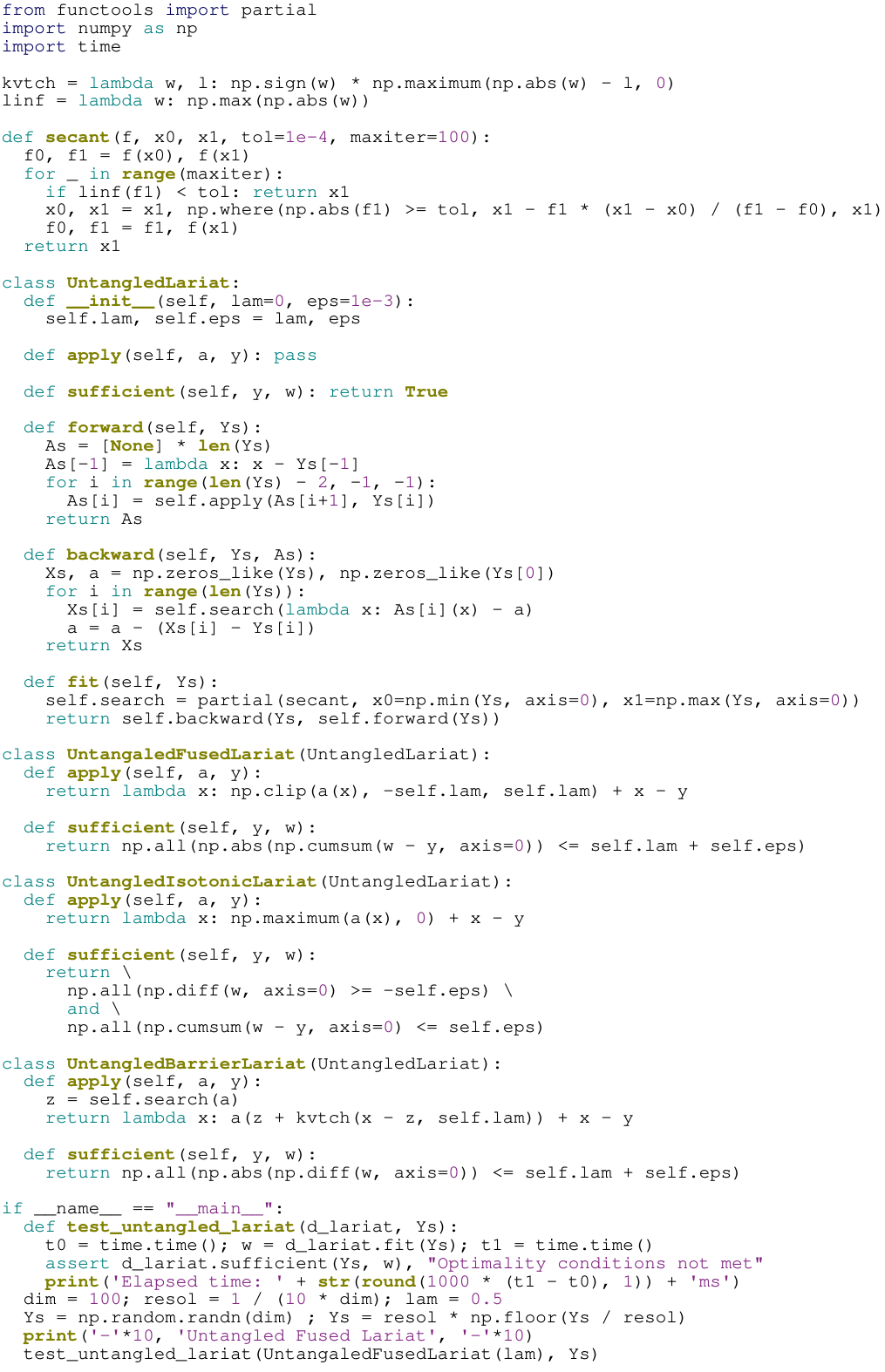}}}}
\caption{Python code of algorithms from Sec.~\ref{sbgf:sec}.}
\end{figure*}

\end{document}